\documentclass[a4paper]{cas-dc}

\usepackage[authoryear, sort&compress]{natbib}

\usepackage{amsmath}
\usepackage{amsfonts}
\usepackage{amssymb}
\usepackage{graphicx}



\def\tsc#1{\csdef{#1}{\textsc{\lowercase{#1}}\xspace}}
\tsc{WGM}
\tsc{QE}

\begin{document}
\let\WriteBookmarks\relax
\def\floatpagepagefraction{1}
\def\textpagefraction{.001}

\shorttitle{}    

\shortauthors{}  

\title [mode = title]{SEG-SAM: Semantic-guided SAM for unified medical image segmentation}  



%



\author[1,2]{Shuangping Huang}
\credit{}

\author[1]{Hao Liang}
\credit{}

\author[1]{Qingfeng Wang}
\credit{}

\author[1]{Chulong Zhong}
\credit{}

\author[1]{Gang Wei}
\credit{}

\author[3]{Miaojing Shi}
\cormark[1]
\credit{}

\affiliation[1]{organization={School of Electronic and Information Engineering, South China University of Technology},
            city={Guangzhou},
            postcode={510641}, 
            country={China}}

\affiliation[2]{organization={Pazhou Lab},
            city={Guangzhou},
            postcode={511442}, 
            country={China}}

\affiliation[3]{organization={College of Electronic and Information Engineering, Tongji University},
            city={Shanghai},
            postcode={200092}, 
            country={China}}

\cortext[1]{Corresponding author at: Tongji University, Shanghai, China. E-mail address: mshi@tongji.edu.cn (M.J. Shi).}



\begin{abstract}
Recently, developing unified medical image segmentation models has gained increasing attention, especially with the advent of the Segment Anything Model (SAM).
SAM has shown promising binary segmentation performance in natural images but still lacks semantic information.
Although subsequent variants expand this paradigm, such as SAM2 and SAM3, they are still restricted to binary segmentation.
Transferring SAM to the medical domain poses further challenges, as medical images often exhibit substantial inter-category overlaps.
In this work, we propose \textbf{SE}mantic-\textbf{G}uided \textbf{SAM} (\textbf{SEG-SAM}), a unified medical segmentation framework that incorporates semantic learning with medical knowledge.
To bridge the gap by transferring SAM into the medical domain while preserving its generalizability, we develop a Semantic-Aware Decoder (SAWD) operating independently of SAM's original decoder.
SAWD follows a dual-flow architecture: 1) a primary flow performs prompted semantic mask prediction via a spatial-semantic disentangled strategy that decouples spatial localization from semantic category assignment, and 2) an auxiliary flow performs potential object classification to derive object prototypes, which are injected into the primary flow for contextual enhancement.
To further enrich semantic understanding, we propose a Text-to-Visual Semantic Enhancement (T2VSE) scheme that incorporates medical text descriptions via a coarse-to-fine process, pruning semantically irrelevant text tokens for fine-grained text-to-vision interaction.
Finally, we introduce spatial alignment and inter-token diversity constraints to promote stable optimization.
Extensive experiments demonstrate that SEG-SAM outperforms state-of-the-art SAM-based methods in unified binary medical segmentation and task-specific methods in semantic medical segmentation, showcasing promising results and potential for broader medical applications.
\end{abstract}





\maketitle

\section{Introduction}\label{sec:intro}
Medical image segmentation aims to delineate regions of interest in medical images, playing a vital role in medical analysis and clinical diagnosis \citep{tmisemiseg, medanddiag11, medanddiag12, medanddiag13, textpromptable, switchumamba}. The segmentation tasks may involve various image modalities (\textit{e.g.}, CT, MRI, and X-rays) and anatomical structures (\textit{e.g.}, organs and tissues). This diversity makes developing task-specific models for each scenario both complex and labor-intensive. Therefore, effectively handling medical segmentation in a unified manner is gaining increasing attention.

To perform unified medical segmentation, a common practice is to train a deep neural network on various medical data \citep{universeg, uniseg, medicaldatamodality}. 
However, this requires carefully designed modules to address the inherent differences across data sources and inter-modal heterogeneity. 
Recently, the Segment Anything Model (SAM) \citep{sam} has shown promising segmentation performance in the natural domain using specifically crafted interactive prompts (\textit{e.g.}, points, bounding boxes) as guidance, paving the way for unified medical segmentation.
Building on this success, its subsequent variants expand this promptable paradigm: SAM2 \citep{sam2} introduces a memory attention mechanism for video segmentation, and SAM3 \citep{sam3} integrates joint vision-language learning for caption-driven segmentation.
Among medical follow-up methods, MedSAM \citep{medsam} applies a standard fine-tuning of SAM on medical data to transfer the pre-trained knowledge from the natural domain into the medical domain. 
Following this, Med2D \citep{sammed2d} and SAM-SP \citep{samsp} use parameter-efficient fine-tuning and knowledge distillation techniques (\textit{e.g.}, adapter \citep{adapter}) to inject the model with medical knowledge.
However, like the original SAM, the interactive prompts in these methods focus strictly on concrete objects for binary segmentation while the semantic attributes (\textit{e.g.}, object categories) are overlooked; even SAM3 remains a binary verification of semantic existence.
Category-specific semantics are crucial for unified medical segmentation, as medical images often possess greater inter-category overlap than natural images. For instance, in order to accurately distinguish between the liver and kidney, which are visually alike and spatially adjacent, we would need not only point-based prompts but also the semantic nuances distinguishing them.
A few recent studies \citep{sammask, medsa} have attempted to integrate semantic information into SAM for medical segmentation.
However, they are restricted to single-modal datasets (\textit{e.g.}, MRI or CT), thus significantly degrading the core nature of SAM as a general-purpose segmentation model.
Adapting SAM for unified semantic medical segmentation requires joint training across multi-modal medical images, which is much more challenging and remains unexplored.

\begin{figure}
\begin{center}
\includegraphics[width=1.0\linewidth]{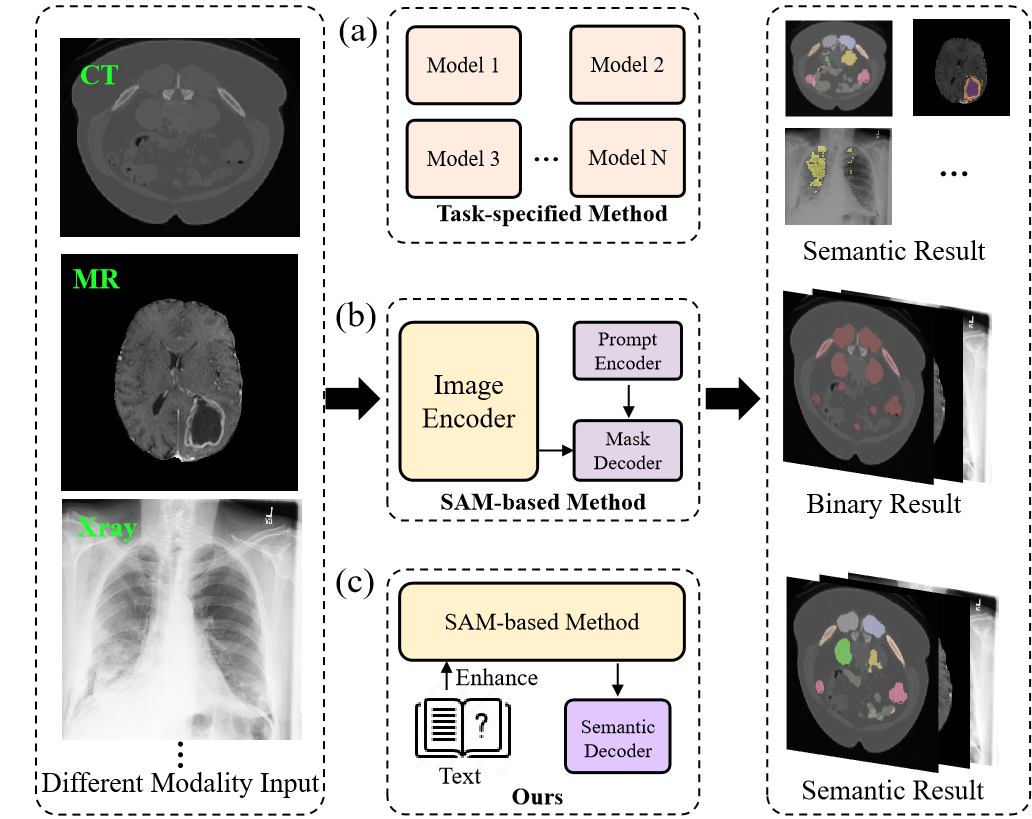}
\end{center}
\caption{Comparison of model architectures for medical segmentation in multi-modal images: (a) traditional methods employ task-specific models that process each image modality independently; (b) SAM-based methods handle various modalities in a unified manner but lack semantic information; (c) our method develops the novel semantic-aware decoder and integrates the textual medical knowledge into the model to enable unified semantic medical segmentation.
}
\label{fig:figure1}
\end{figure}

In this paper, we aim to incorporate semantic learning into SAM to facilitate unified medical segmentation. It is, however, not straightforward to fine-tune the pre-trained SAM with semantic medical labels, owing to the performance degradation caused by the classifier weight distribution shift from binary to semantic segmentation (see Fig. \ref{fig:figure_kde}).
To address this, we develop a \textbf{semantic-aware decoder} (SAWD), independent from SAM's original semantic-agnostic decoder (SAGD), which is specialized in learning semantic knowledge of medical categories.
We construct SAWD by dual flows: a primary flow dedicated to semantic mask prediction for the prompted object and an auxiliary flow that focuses on the potential object context for semantic enhancement.
For the primary flow, we employ a similar design to SAGD equipped with a special \textit{segmentation-oriented token} (SO-Token).
Guided by visual prompts, the SO-Token interacts with visual features to obtain an object representation. 
In order to achieve semantic prediction yet without sacrificing  SAM's zero-shot generalizability, we propose a \textit{spatial-semantic disentangled mask prediction} strategy. 
This design enables the SO-Token to separately drive spatial localization and semantic category assignment, thereby allowing our framework to naturally adapt to novel categories via similarity-based assignment.
For the auxiliary flow, we employ \textit{classification-oriented tokens} (CO-Tokens) for potential category prediction without prompts, from which we derive \textit{object prototypes} and inject them into the SO-Token to provide rich contextual knowledge.
To enrich the model's semantic understanding, we leverage the large language model (LLM) to generate and encode medical category descriptions, and then integrate them into our model, inspired by text promptable segmentation \citep{sam, textpromptable}.
We propose the \textbf{text-to-vision semantic enhancement} (T2VSE) scheme to effectively incorporate this knowledge.
Specifically, T2VSE adopts a coarse-to-fine interaction process, where \textit{coarse text token selection} is first applied to prune semantically irrelevant text tokens, followed by \textit{fine text-to-vision interaction} within the retained token subset to summarize medical semantic knowledge for accurate segmentation.

To ensure stable model optimization, we introduce two auxiliary constraint losses tailored to SAWD and T2VSE. First, a \textit{cross-mask spatial alignment loss} is designed to enforce mask consistency between the independent predictions of the SAWD and SAGD.
Second, an \textit{inter-token diversity loss} is employed to explicitly encourage the text summary tokens in T2VSE to capture distinct representation patterns.
Owing to their highly modular design, our proposed SAWD and T2VSE can be seamlessly incorporated into advanced SAM variants, including SAM2 and SAM3, with minimal modifications.

To the best of our knowledge, we are the first to introduce semantic learning into SAM to achieve \textbf{unified multi-modal medical semantic segmentation}.
We denote the proposed model as \textbf{SE}mantic-\textbf{G}uided \textbf{SAM} (\textbf{SEG-SAM}) and Fig. \ref{fig:figure1} illustrates the distinct differences between our method and previous studies.
Extensive experiments on various publicly available datasets demonstrate that the proposed SEG-SAM outperforms other SAM-based models on binary medical segmentation, while offering additional benefits on semantic prediction. Additionally, we evaluate the medical semantic segmentation performance between our method and state-of-the-art methods, showcasing promising results and the potential for wider medical application. Importantly, while extending SAM with semantic learning capability, our architecture effectively preserves its inherent zero-shot generalization, as also substantiated by our experimental evaluation.

\begin{figure*}
\begin{center}
\includegraphics[width=0.95\linewidth]{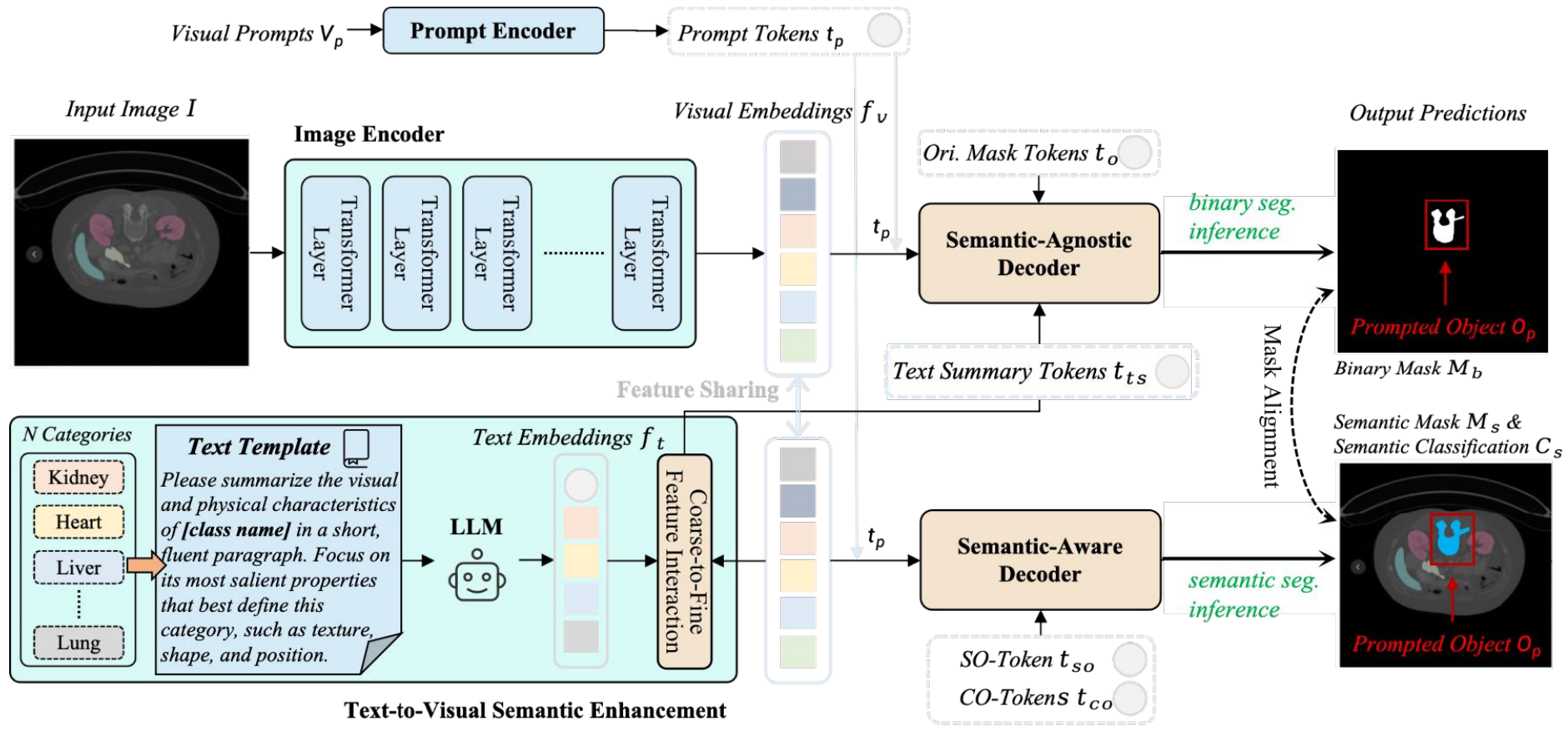}
\end{center}
\caption{Overview of the SEG-SAM framework. Given an input image $I$ and a prompted object $O_p$ from visual prompts $V_p$, first, SAM's image encoder extracts visual embeddings $f_v$, while the prompt encoder encodes visual prompts $V_p$ into prompt tokens $t_p$. Next, SAM's original decoder, namely semantic-agnostic decoder (SAGD), employs the original mask tokens $t_o$ to predict a binary mask $M_b$ for $O_p$. Then, our proposed semantic-aware decoder (SAWD) utilizes a segmentation-oriented token (SO-Token) $t_{so}$ to predict a semantic mask $M_s$ for $O_p$, and classification-oriented tokens (CO-Tokens) $t{co}$ to perform semantic classification $C_s$ of potential objects. Meanwhile, medical text descriptions generated from an LLM using a text template are incorporated into the text summary tokens $t_{ts}$ through a text-to-vision semantic enhancement scheme, and $t_{ts}$ is integrated into SAGD for segmentation enhancement. 
During training, we use mask alignment and inter-token diversity of $t_{ts}$ constraints to ensure stable optimization.
}
\label{fig:figure2}
\end{figure*}

\section{Related works}\label{sec:relatedworks}
\subsection{Medical image segmentation}\label{sec:medimgseg}
Driven by deep neural networks (DNNs), medical image segmentation has achieved significant progress. 
The FCN \citep{fcn} pioneers a fully convolutional architecture for end-to-end image segmentation. Following this, U-Net \citep{unet} stands as one of the most successful medical image segmentation methods, leveraging skip connections between encoder and decoder to capture multi-scale contextual features. Later, U-Net variants \citep{unet++, unet3+, denseunet} further enhance robustness by integrating components like adaptive re-sampling and hierarchical feature extraction modules, addressing different medical challenges. 
The recent success of Transformer architecture \citep{transformer,vit, swin} in computer vision has also inspired hybrid designs \citep{swinunet, transunet}, \textit{i.e.}, combining attention mechanisms and U-shaped networks, for global-local feature extraction. However, despite their excellent performance, all these methods are task-specific and cannot generalizable across tasks.

Recently, an increasing number of studies have aimed at unified semantic medical segmentation, which handles multiple medical datasets within a single architecture \citep{universeg, universalllm, uniseg}. 
UniSeg \citep{uniseg} trains the model on aggregated datasets to learn inter-dataset correlations, enhancing its segmentation capability.
UniverSeg \citep{universeg} employs cross-dataset image pairs to build a unified model that is able to generalize to new tasks without fine-tuning.
HermesNet \citep{hermesnet} reduces the heterogeneity in multi-modal medical images by jointly modeling task and modality priors for medical segmentation.
Although these methods build unified models, they require explicit architectural designs to mitigate domain gaps across datasets, until the introduction of Segment Anything Model (SAM) \citep{sam}.

\subsection{SAM in medical image segmentation}
SAM series models represent a foundational paradigm shift for promptable segmentation.
The original SAM leverages interactive visual prompts to perform binary mask prediction.
SAM2 \citep{sam2} extends the framework to video segmentation via memory attention and upgrades with a multi-scale ViT encoder for richer feature extraction.
More recently, SAM3 \citep{sam3} advances the paradigm into caption-driven segmentation through joint vision-language representation learning, incorporating a pretrained text encoder and a transformer-based detector module.
Despite supporting text phrase-based prompting, SAM3's core mechanism still predominantly performs binary verification of category presence rather than explicit semantic prediction, and it struggles to effectively comprehend the long, detailed text descriptions.

To bridge the gap between these general foundation models and specialized applications, subsequent studies have extended the SAM framework to diverse domains \citep{samnerf, samhq, samfromspace, samrobust, glandseg}.
Among these efforts, applying SAM to the medical domain often requires model fine-tuning on medical images due to the significant domain gap.
To achieve this, MedSAM \citep{medsam} and SAM-Med2D \citep{sammed2d} introduce large-scale medical datasets to fine-tune SAM and adapt its knowledge to the medical segmentation task.
Furthermore, SAM-Adapter \citep{samadapter} incorporates the high-frequency components of medical images into a lightweight adapter as enhanced visual signals, thereby integrating medical-specific knowledge into SAM through fine-tuning.
SAM-SP \citep{samsp} introduces a self-prompting scheme to improve object mask prediction and further enhances the segmentation performance through knowledge distillation between generated masks.
Despite these advancements, existing SAM-based methods remain confined to binary segmentation due to the limitation of the prompting architecture. 
Following this trajectory, while SAM2-Adapter \citep{sam2adapter} and SAM3-Unet \citep{sam3unet} fine-tune SAM's variants on specific medical datasets, recent SAM3 adaptations, including Medical-SAM3 \citep{medicalsam3}, MedSAM3 \citep{medsam3}, and AdaSAM3 \citep{adasam3}, build multi-modality medical datasets configured with object-phrase-based captions to achieve joint text-visual prompt fine-tuning.
Although these methods preserve SAM3's text prompt-based semantic representation learning, they still lack a dedicated mechanism for explicit semantic class prediction.
Recently, MaskSAM \citep{sammask} and Med-SA \citep{medsa} attempt to incorporate semantic learning into SAM by introducing adapters to directly fine-tune the original decoder. While conferring a degree of explicit semantic prediction, these methods are limited to single-modal images and suffer from substantial performance degradation as they directly extend the binary decoder for semantic prediction.

\section{Methodology}
We begin by reviewing the architecture of SAM series models; then we introduce the SEG-SAM for unified medical image segmentation; and lastly, we describe the loss function, training, and inference strategies of SEG-SAM.

\subsection{Preliminary: SAM series models}\label{sec: preliminary}
The original \textbf{SAM} \citep{sam} primarily consists of three components: 1) an image encoder $\mathrm{Enc_{img}}$ powered by a vision transformer (ViT) \citep{vit} to extract visual embeddings $f_v$ from the input image $I$; 2) a prompt encoder $\mathrm{Enc_{pmt}}$ that encodes visual prompts $V_p$ into prompt tokens $t_p$; and 3) a semantic-agnostic decoder $\mathrm{Dec_{sag}}$ that facilitates interaction among $f_v$, $t_p$, and learnable tokens $t_o$. The refined visual features are subsequently mapped to generate the final binary masks $M_s$. 
As architectural variants, \textbf{SAM2} \citep{sam2} (in its static-image configuration) retains SAM's core pipeline but upgrades $\mathrm{Enc_{img}}$ into a hierarchical ViT for multi-scale feature extraction.
\textbf{SAM3} \citep{sam3} further augments the architecture for caption-driven segmentation by incorporating two additional components: 4) a text encoder $\mathrm{Enc_{txt}}$ to extract text prompts from category phrases, and 5) a transformer-based detector $\mathrm{Det}$ for vision-text fusion and final mask prediction.
Without loss of generality, we instantiate our proposed algorithm based on SAM and detail its extensions to SAM2 and SAM3 later.

\begin{figure}
\begin{center}
\includegraphics[width=0.95\linewidth]{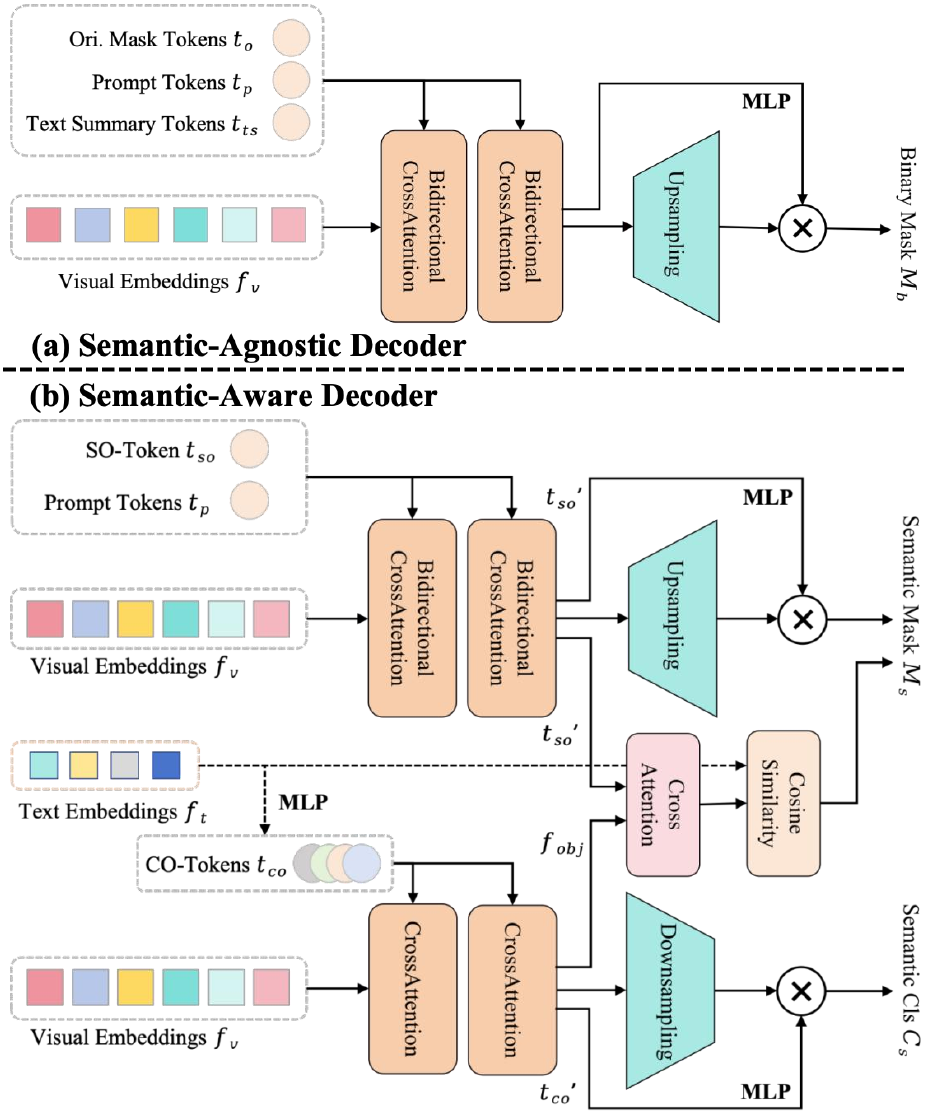}
\end{center}
\caption{Architectural details of (a) the semantic-agnostic decoder in SAM and (b) the proposed semantic-aware decoder.}
\label{fig:figure3}
\end{figure}

\subsection{Semantic-guided SAM}
\label{sec:overall}
The proposed SEG-SAM framework preserves SAM's original design to maintain its representation learning, while incorporating additional semantic learning. As shown in Fig.~\ref{fig:figure2}, given an image $I \in \mathbb{R}^{3 \times H \times W}$ and a prompted object $O_p$ indicated by visual prompts $V_p$ (\textit{i.e.}, points and bounding boxes),
we adopt SAM's image encoder $\mathrm{Enc_{img}}$ to extract visual embeddings $f_v \in \mathbb{R}^{D \times H/p \times W/p}$ and its prompt encoder $\mathrm{Enc_{pmt}}$ to encode $V_p$ into prompt tokens $t_p \in \mathbb{R}^{L \times D}$, where $D$ denotes the feature dimension, $p$ is the patch size, and $L$ is the number of point prompts (a bounding box can be treated as two corner points), specified below:
\begin{equation}
\label{meth:Eq.1}
f_{v} = \mathrm{Enc_{img}}(I),\ t_p = \mathrm{Enc_{pmt}}(V_p).
\end{equation}
\noindent Given $f_v$ and $t_p$, we retain SAM's semantic-agnostic decoder $\mathrm{Dec_{sag}}$ and generate original mask tokens $t_o$. These tokens are then reused as inputs to $\mathrm{Dec_{sag}}$ to predict a binary mask $M_b$ for the prompted object $O_p$, as shown in Fig.~\ref{fig:figure3}(a).
To introduce semantic learning capabilities, 
we develop a \textbf{semantic-aware decoder} $\mathrm{Dec_{saw}}$ that generates two types of tokens:
1) the \textit{segmentation-oriented token} (SO-Token), denoted as $t_{so}$, for predicting the semantic mask $M_s$ of $O_p$, and 2) the \textit{classification-oriented tokens} (CO-Tokens), denoted as $t_{co}$, for predicting the semantic categories $C_s$ of all objects in the image. 
Through iterative refinement within $\mathrm{Dec_{saw}}$, $t_{co}$ provide auxiliary guidance to $t_{so}$, enabling accurate semantic mask prediction (Sec.~\ref{sec:SAD}).
To further enhance semantic understanding, we solicit medical text descriptions $D_{text}$ and encode them as text embeddings $f_t$. Subsequently, we propose the \textbf{text-to-vision semantic enhancement} (T2VSE) module, which enables the coarse-to-fine feature interaction between $f_t$ and $f_v$ (followed by prompts $t_p$) and summarizes medical knowledge through text summary tokens $t_{ts}$. We then inject $t_{ts}$ into $\mathrm{Dec_{sag}}$ for segmentation refinement (Sec.~\ref{sec:Text}). The overall process can be specified below:
\begin{equation}
\label{meth:Eq.2}
f_{t}=\mathrm{LLM\_Embed}(D_{text}),\ t_{ts} = \mathrm{T2VSE}(f_t, f_v, t_p),
\end{equation}
\begin{equation}
\label{meth:Eq.3}
M_b = \mathrm{\mathrm{Dec_{sag}}}([t_{o}, t_{ts}, t_p], f_{v}),
\end{equation}
\begin{equation}
\label{meth:Eq.4}
M_s, C_{s} = \mathrm{\mathrm{Dec_{saw}}}([t_{so}, t_{co}, t_{p}], f_{v}).
\end{equation}
\noindent Next, we detail two key components of the proposed SEG-SAM, \textit{i.e.}, the semantic-aware decoder and text-to-vision semantic enhancement scheme.

\begin{figure}
\begin{center}
\includegraphics[width=1.0\linewidth]{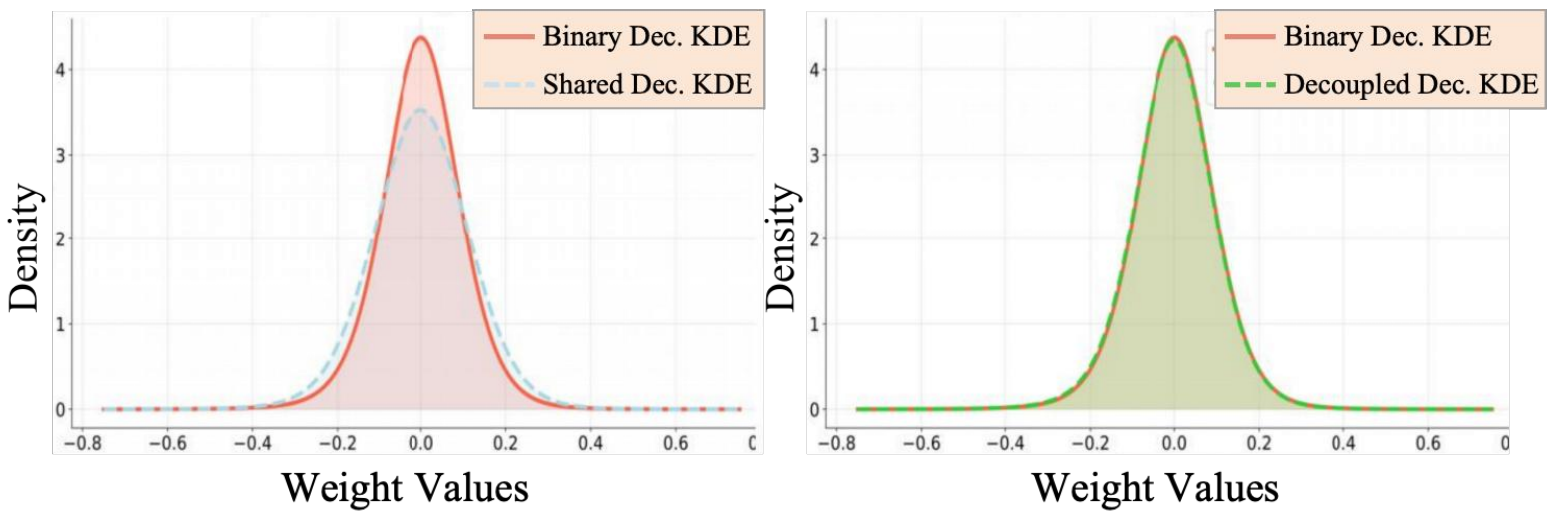}
\end{center}
\caption{Visualization of weight distributions of classifiers from binary to semantic segmentation: The decoder sharing (left) alters the weight distribution of the original classifier, whereas the decoder decoupling (right) retains it.
}
\label{fig:figure_kde}
\end{figure}

\begin{figure}
\begin{center}
\includegraphics[width=0.95\linewidth]{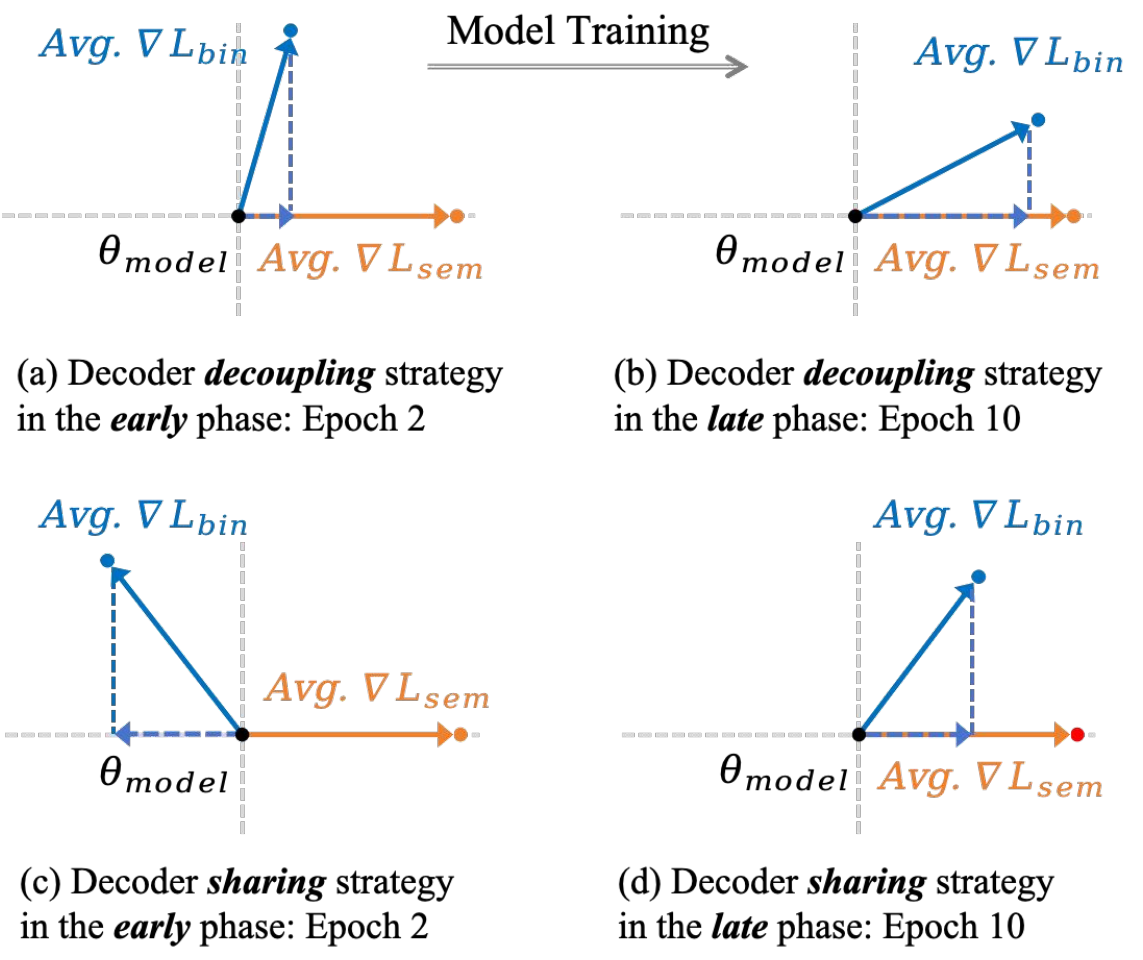}
\end{center}
\caption{Visualization of average gradient directions derived from binary and semantic segmentation losses. 
The gradient conflict is amplified from the decoder sharing to the decoder decoupling, and gradually diminishes as training progresses.
}
\label{fig:figure_grad}
\end{figure}

\subsubsection{Semantic-aware decoder}
\label{sec:SAD}
In our setup, SAM is first pre-trained on the natural binary segmentation task and then jointly fine-tuned on both binary and semantic medical segmentation tasks. 
For joint learning, we develop two possible strategies: one is the \textit{decoder sharing strategy}, which uses a shared decoder followed by two independent segmentation heads; the other is the \textit{decoder decoupling strategy}, which assigns each learning task its own dedicated decoder along with a prediction head. Below we discuss the two strategies and choose the latter in the end.

During the model training process, the gradient descent-based optimizer adopts momentum-based weights update. Let $w_t$ and $\nabla f(w_t)$ be respectively the weight and gradient at iteration $t$, and $\alpha$ be the step size. The momentum-based update rule is given by the following:
\begin{equation}
\label{meth:Eq.5}
w_t=w_{t-1}-z_t;\ z_t=\beta~z_{t-1}+\alpha~\nabla~f(w_{t-1}),
\end{equation}
\noindent where $z$ is the momentum and $\beta$ is the momentum parameter. The momentum effectively acts as a moving average of gradients, smoothing oscillations.

Unlike the binary segmentation task, which emphasizes foreground-background distinctions, the semantic segmentation task focuses on the extraction of fine-grained, class-specific features. 
When a shared decoder is used (decoder sharing strategy), the gradients backpropagated from the binary and semantic losses must share the same momentum in the decoder; yet, because of the inconsistent focus between the two tasks, their gradients can be incompatible, sometimes with contradictory directions.
Using one momentum to record both histories will make the network weight optimization toggle between the two sources of gradients. This can lead to adversarial effects for the learning of both tasks, ending up with oscillation and slow convergence. 
In contrast, by employing our decoder decoupling strategy, each task has its own decoder parameters and independent momentum. This avoids gradient conflict in the shared decoder and can thus end up with better and faster convergence.
Fig.~\ref{fig:figure_kde} shows that \textit{this design can retain the weight distribution of SAM's binary classifier, whereas the decoder sharing strategy leads to a clear weight distribution shift, severely degrading the segmentation performance originally equipped in SAM.}
We further measure the respective gradient for the binary and semantic segmentation tasks backpropagated through the shared decoder or decoupled decoders to the visual encoder: when examining a specific epoch, the gradient conflict between the two segmentation tasks is clearly amplified from the decoder decoupling strategy (Fig.~\ref{fig:figure_grad}a) to the decoder sharing strategy (Fig.~\ref{fig:figure_grad}c).
Furthermore, as shown in Fig.~\ref{fig:figure_grad}b and Fig.~\ref{fig:figure_grad}d, gradient conflicts for both strategies diminish with an increase of training epochs, since binary and semantic segmentation indeed segment the same object and should eventually converge to a similar performance level; \textit{throughout this process, our decoder decoupling strategy converges faster with less gradient conflict and achieves a relatively higher performance level at the end} (see Fig.~\ref{fig:figure_loss}).

To implement this decoder decoupling strategy, we develop a dedicated semantic-aware decoder $\mathrm{Dec_{saw}}$ for semantic learning while retaining SAM's original binary decoder.
Specifically, $\mathrm{Dec_{saw}}$ incorporates two learning flows to simultaneously perform semantic segmentation of the prompted object and classification of all objects in the image.
This design facilitates precise object segmentation that is enriched by context from the other objects, while avoiding the computational cost of extensively segmenting every object.
These two learning flows are detailed as follows.

\noindent \textbf{Semantic mask prediction for the prompted object.}
\label{prompted tokens}
We begin with semantic label generation by defining the ground-truth (GT) semantic masks $\mathcal{M}_\mathrm{sem}$ to replace the original binary masks $\mathcal{M}_\mathrm{bin}$ in the unified dataset.
Specifically, each GT binary mask $M \in \mathcal{M}_\mathrm{bin}$ is associated with a semantic embedding $f_t^M$, which is encoded from the corresponding object text description generated by an LLM (detailed in Sec. \ref{sec:Text}).
Based on this matching, we formulate the GT semantic masks directly as mask-embedding pairs, denoted as $\mathcal{M}_\mathrm{sem}=\{(M,f_t^M)|M \in \mathcal{M}_\mathrm{bin}\}$.
Next, to facilitate the token-vision feature learning, we initialize a learnable SO-Token $t_{so}$ within $\mathrm{Dec_{saw}}$, driving the subsequent interaction.
Given the prompt tokens $t_p$ from $\mathrm{Enc_{pmt}}$ and the visual embeddings $f_v$ from $\mathrm{Enc_{img}}$,
we apply two bidirectional attention layers to perform feature interactions. In each layer, $t_{so}$ is concatenated with $t_p$ and processed by a self-attention operation for token interaction. Then, a bidirectional cross-attention operation is employed to refine both the concatenated tokens and $f_v$, yielding the updated tokens $t_{so}'$ and visual embeddings $f_v'$.
Finally, to obtain the semantic mask $M_s$ of the prompted object $O_p$, we introduce the spatial-semantic disentangled mask prediction strategy.

\noindent \textit{Spatial-semantic disentangled mask prediction.} This strategy decouples semantic mask prediction into \textit{mask generation} and \textit{semantic matching}. Specifically, the spatial mask is generated via a dot product between the refined tokens $t_{so}'$ and the visual embeddings $f_v'$. Concurrently, the semantic matching score is derived by calculating the cosine similarity between $t_{so}'$ and the semantic embeddings $f_t$. 
The overall process is defined as:
\begin{equation}
\label{meth:Eq.5}
t_{so}', f_v' = \mathrm{BiCrossAttn}(\mathrm{SelfAttn}([t_{so},t_p]), f_v),
\end{equation}
\begin{equation}
\label{meth:semmask}
M_s = (\underbrace{\mathrm{MLP}(t_{so}') \times \mathrm{DC}(f_v')}_{\text{Mask generation}},\ \underbrace{\mathrm{CosSim}(\mathrm{Lin}(t_{so}'), f_t)}_{\text{Semantic matching}}),
\end{equation}
\noindent where $\mathrm{DC}$ is the deconvolutional layer for upsampling and $\mathrm{CosSim}$ is the cosine similarity function.
By doing this, the SO-Token separately processes spatial localization and semantic category assignment of the prompted object, naturally facilitating semantic matching for novel categories.

\noindent \textbf{Semantic classification for all objects in the image.}
Intuitively, the semantic learning of other objects in the image can provide the category-specific context to further facilitate the semantic segmentation on the prompted object.
To this end, we introduce CO-Tokens $t_{co}$, each associated with a candidate object category, and initialize them as \textit{text-conditioned tokens} for potential category prediction.
Instead of random initialization, we construct $N$ CO-Tokens by projecting predefined object semantic embeddings $f_t$ via an MLP. 
$N$ is determined by the number of text descriptions used to obtain $f_t$ (detailed in Sec. \ref{sec:Text}) and can be set equal to or greater than the training category size.
Then, to perform token-vision feature interaction, $t_{co}$ are processed by two attention layers in $\mathrm{Dec_{saw}}$ to interact with the visual embeddings $f_v$.
Each layer applies sequential self-attention ($Q,K,V=t_{co}$) and cross-attention ($Q=t_{co}; K,V=f_v$) to refine the tokens, yielding $t_{co}'$.
The final classification logits $C_s$ are obtained by computing the cosine similarity between the refined tokens $t_{co}'$ and the visual features $f_v$, as follows:
\begin{equation}
\label{meth:Eq.8}
t_{co}' = \mathrm{CrossAttn}(\mathrm{SelfAttn}(t_{co}), f_v),
\end{equation}
\begin{equation}
\label{math:multicls}
C_s = \mathrm{CosSim}(\mathrm{MLP}(t_{co}') \times \mathrm{DS}(\mathrm{Conv}(f_v))),
\end{equation}
where $\mathrm{DS}$ is the downsampling function.
To improve semantic mask prediction, we subsequently leverage $t_{co}'$ to construct category-specific object prototypes $f_{obj}$.

\noindent \textit{Category-specific object prototypes.} The prototypes serve as contextual knowledge for the refined SO-Token $t_{so}'$, enriching its representation to facilitate semantic matching.
Specifically, we compute the spatial similarity matrix between $t_{co}'$ and $f_v$, and then apply the softmax function $\sigma$ to derive attention weights for aggregating $f_v$ to construct $f_{obj}$, as formulated below:
\begin{equation}
\label{meth:Eq.9}
f_{obj} = \sigma(\mathrm{CosSim}(\mathrm{MLP}(t_{co}') \times \mathrm{Conv}(f_v))) \times f_v.
\end{equation}
\noindent Based on this, we inject $f_{obj}$ into $t_{so}'$ via a cross-attention operation, updating the formulation of semantic matching in Eq. \ref{meth:semmask} to: $\mathrm{CosSim}(\mathrm{Lin}(\mathrm{CA}(t_{so}',f_{obj})), f_t)$. This approach eliminates the need to explicitly segment all potential objects, reducing computational cost while obtaining superior performance.

\begin{figure}
\begin{center}
\includegraphics[width=1.0\linewidth]{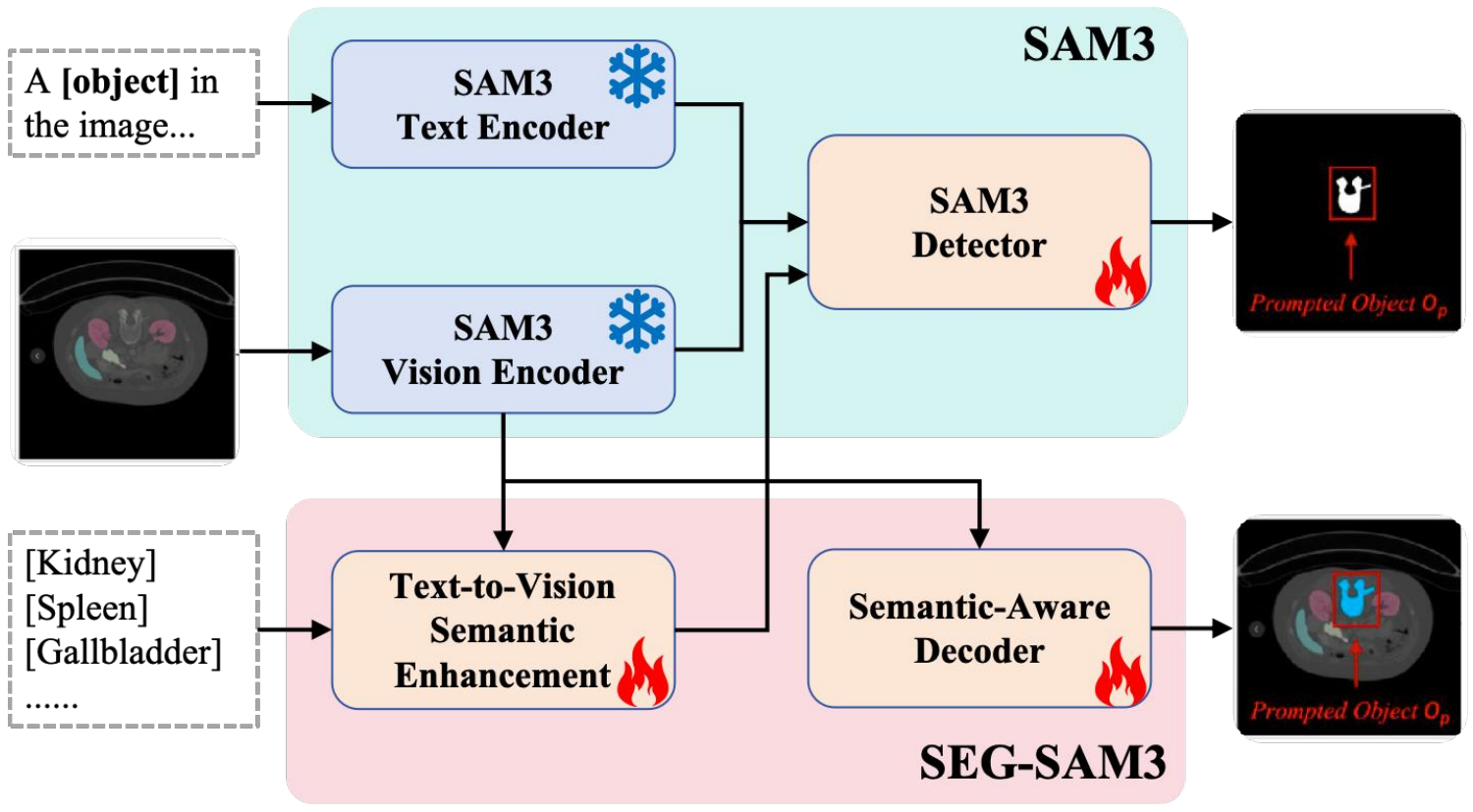}
\end{center}
\caption{Architectural details of the proposed SEG-SAM3, built upon the original SAM3 framework.}
\label{fig:figure_sam3}
\end{figure}

\subsubsection{Text-to-vision semantic enhancement}
\label{sec:Text}
To enhance fine-grained semantic understanding, we propose a \textit{text-to-vision semantic enhancement} scheme that incorporates text descriptions detailing key characteristics of object categories.
We first present the generation and encoding of these descriptions, followed by a coarse-to-fine cross-modal interaction that aggregates medical semantic knowledge and injects it into the semantic-agnostic decoder $\mathrm{Dec_{sag}}$ to guide mask prediction.

\noindent \textbf{Text description generation and encoding.}
To acquire these informative descriptions, we first utilize a pretrained LLM, \textit{i.e.}, ChatGPT \citep{chatgpt}, to generate the text descriptions $D_{text}$ of global object categories.
We construct a generic text template applicable to any given category, guiding the LLM to extract the most salient visual properties, encompassing aspects like texture, shape, and position.
The complete text template is as follows: 
\textit{Please summarize the visual and physical characteristics of [class name] in a short, fluent paragraph. Focus on its most salient properties that best define this category, such as texture, shape, and position.}
The generated $D_{text}$ are then manually curated by two expert clinicians to eliminate potential inaccuracies.
Once verified descriptions $D_{text}$ for global object categories are obtained, we encode them into text embeddings $f_{t} \in \mathbb{R}^{N \times D}$ using a GPT-embedding model, where $N$ denotes the number of object categories. These embeddings serve as the foundation for our proposed text-to-vision semantic enhancement scheme.

\begin{table*}[t]
\begin{center}
    \footnotesize
    \caption{Comparison between our method, conventional methods, and state-of-the-art SAM-based methods on the Med2D-16M dataset. We use the point and bounding box as the visual prompts for SAM-based methods, and use the DSC metric to evaluate the binary medical segmentation performance across 8 modalities, and additionally report the average DSC over all modalities.}
    \label{tab:bin_seg}
    \begin{tabular}{l|c|cccccccc|c}
        \toprule
        Method & Prompt & CT & MR & PET & Derm. & Endo.  & Fund. & Ultr. & X-ray & Avg. \\
        \midrule
        SAM \citep{sam} & \multirow{7}{*}{Point} &  29.03 &	22.55&	35.43&	35.74&	31.61&	30.34&	18.88&	19.79&	27.92\\
        FT-SAM \citep{sammed2d} & & 63.58&	41.72&	68.24&	82.55&	58.02&	71.28&	62.94&	44.70&	61.63\\
        FT-SAM2 \citep{sam2} & & 62.19&	43.81&	71.66&	83.05&	57.22&	73.31&	63.90&	45.11&	62.53 \\
        SAM-Med2D \citep{sammed2d} & &  73.63&	50.52&	75.11&	87.59&	60.23&	76.16&	69.25&	53.28&	68.22\\
        MedSAM \citep{medsam} & & 74.31&	52.08&	72.70&	88.02&	58.47&	76.78&	70.33&	55.13&	68.48 \\
        Med-SA \citep{medsa} & &  73.82&	49.87&	75.35&	88.64&	60.79&	77.17&	69.51&	48.43&	67.95 \\
        \textbf{SEG-SAM (Ours)} & & \textbf{78.76} & \textbf{53.68} & \textbf{75.55} & \textbf{90.38} & \textbf{71.04} & \textbf{85.18} & \textbf{71.23} & \textbf{61.05} & \textbf{73.36} \\
        \textbf{SEG-SAM2 (Ours)} & & \textbf{77.39} & \textbf{54.37} & \textbf{75.91} & \textbf{90.24} & \textbf{70.55} & \textbf{85.53} & \textbf{72.02} & \textbf{61.83} & \textbf{73.48} \\
        \midrule
        SAM \citep{sam} & \multirow{7}{*}{Box} & 64.29&	51.34&	64.73&	85.27&	77.44&	77.41&	77.38&	52.39&	68.78 \\
        FT-SAM \citep{sammed2d} & & 80.41&	60.57&	77.34&	90.85&	78.17&	84.84&	82.25&	60.16&	76.82 \\
        FT-SAM2 \citep{sam2} & & 80.13&	60.26&	78.03&	91.22&	78.64&	85.08&	82.52&	60.93&	77.10 \\
        SAM-Med2D \citep{sammed2d} & & 83.49&	62.38&	78.71&	93.17&	78.38&	87.27&	84.64&	59.49&	78.44 \\
        MedSAM \citep{medsam} & & 84.28&	63.84&	77.32&	91.75&	76.13&	85.38&	84.47&	65.92&	78.64 \\
        Med-SA \citep{medsa} & & 84.66&	63.90&	80.54&	92.38&	79.06&	86.83&	83.59&	57.26&	78.53 \\
        FT-SAM3 \citep{sam3} & & 84.31&	63.76&	80.12&	93.79&	80.46&	84.60&	84.28&	64.71&	79.49 \\
        \textbf{SEG-SAM (Ours)} & & \textbf{86.58} & \textbf{66.45} & \textbf{82.46} & \textbf{94.28} & \textbf{81.96} & \textbf{88.06} & \textbf{86.64} & \textbf{68.37} & \textbf{81.85} \\
        \textbf{SEG-SAM2 (Ours)} & & \textbf{85.79} & \textbf{67.53} & \textbf{82.61} & \textbf{93.90} & \textbf{81.67} & \textbf{88.32} & \textbf{86.81} & \textbf{69.05} & \textbf{81.96} \\
        \textbf{SEG-SAM3 (Ours)} & & \textbf{86.83} & \textbf{68.51} & \textbf{83.09} & \textbf{94.46} & \textbf{82.58} & \textbf{88.34} & \textbf{87.07} & \textbf{70.81} & \textbf{82.71} \\
        \bottomrule
    \end{tabular}
\end{center}
\end{table*}

\noindent \textbf{Coarse-to-fine feature interaction.}
Expert-curated text descriptions provide generic overviews of object categories, applying them directly to certain image may potentially lead to misalignment due to individual image variations on perspectives, occasions, etc.
To address this, we propose a progressive \textit{coarse-to-fine interaction} process.
Specifically, we first propose a coarse text-vision interaction to prune the irrelevant/mismatched tokens from text descriptions to certain object in the medical image. Afterwards, a fine-grained text-to-vision interaction is subsequently applied to selectively aggregate text embeddings and integrate with visual information. 
We detail the concrete mechanisms below.

\noindent \textit{The coarse interaction stage.}
Specifically, we first perform \textit{text token selection} to filter out semantically distant text tokens and suppress spurious noise.
We compute the semantic correlation between spatially pooled visual features $\mathrm{Pool}(f_v)$ and global text embeddings $f_t$, where the pooling operation alleviates the influence of unaligned local spatial details.
Based on the resulting correlation scores, we apply a Top-$K$ operation over the complete set of $N$ category text tokens to retain the $K$ text tokens highly relevant to the visual features, forming the pruned text embedding subset $f_t' \in \mathbb{R}^{K \times D}$. This process is formalized as follows:
\begin{equation}
f_{t}'=\mathrm{TopK}_{f_t}(\mathrm{Sim}(\phi(\mathrm{Pool}(f_v)),\ \phi(f_t))),
\end{equation}
\noindent where $\phi(\cdot)$ denotes the sequential operation of linear projection followed by L2 normalization.

\noindent \textit{The fine interaction stage.}
We then establish fine-grained text-vision interactions to enrich model representations with medical text knowledge.
To compress such text-enhanced representations into a compact set of token features for subsequent decoder guidance, we introduce $M$ learnable text summary tokens $t_{ts} \in \mathbb{R}^{M \times D}$.
Specifically, we first concatenate $t_{ts}$ with $f_t'$ into $[f_t',t_{ts}]$, which serves as query, key, and value within a self-attention operation to facilitate token interaction.
Then, the updated tokens by self-attention are concatenated with prompt tokens $t_p$ to form query, while visual features $f_v$ serve as key and value to perform text-to-vision interaction via a cross-attention operation. 
This process is expressed as:
\begin{equation}
\label{Eq. 9}
t_{ts}' = \mathrm{CrossAttn}([\mathrm{SelfAttn}([f_{t}', t_{ts}]), t_p], f_v).
\end{equation}

\noindent \textit{Knowledge injection.}
Finally, we re-project $t_{ts}'$ through an MLP and integrate them into $\mathrm{Dec_{sag}}$ to guide binary segmentation.
Since $\mathrm{Dec_{sag}}$ inherently lacks category semantics, $t_{ts}$ compensates by injecting enhanced object knowledge.
In contrast, this integration is omitted in $\mathrm{Dec_{saw}}$, where explicit semantic cues from GT labels are already available.
Furthermore, to encourage $t_{ts}$ to capture distinct representational patterns, we introduce an inter-token diversity loss, detailed in Sec. \ref{sec:training}.

\begin{table*}[h]
\begin{center}
    \footnotesize
    \caption{Comparison between our method, conventional methods, and state-of-the-art SAM-based methods on the Med2D-16M dataset. We use the point and bounding box as the visual prompts for SAM-based methods, and use the mDSC metric to evaluate the semantic medical segmentation performance across 8 modalities and report the average mDSC. The "*" symbol indicates that we integrate our semantic-aware decoder into them.}
    \label{tab:sem_seg}
    \begin{tabular}{l|c|cccccccc|c}
        \toprule
        Method & Prompt & CT & MR & PET & Derm. & Endo.  & Fund. & Ultr. & X-ray & Avg. \\
        \midrule
        nnUNet \citep{nnunet} & \multirow{4}{*}{-} & 70.59&	61.36&	64.77&	78.14&	78.21&	9.49&	54.27&	77.44&	61.78 \\   
        UNETR \citep{unetr} &  & 71.46&	63.56&	67.44&	88.16&	80.14&	11.57&	58.60&	80.82&	65.22 \\     
        nnFormer \citep{nnformer} &  & 70.26&	64.07&	68.41&	84.28&	78.86&	40.95&	62.52&	81.19&	68.82 \\   
        U-Mamba \citep{umamba} &  & 71.83&	63.78&	71.28&	85.41&	80.67&	36.64&	62.09	&80.21&	68.98 \\    
        \midrule
        Med2D* \citep{sammed2d} & \multirow{4}{*}{Point} & 68.64& 	48.14& 	68.72& 	85.36& 	69.53& 	70.47& 	71.34& 	81.94& 	70.52 \\
        MedSAM* \citep{medsam} &  & 66.28&	49.57&	70.59&	87.37&	64.21&	82.95&	66.55&	69.11&	69.58 \\
        Med-SA* \citep{medsa} &  & 67.52&	48.29&	70.15&	89.03&	65.62&	86.18&	74.28&	82.34&	72.93 \\
        \textbf{SEG-SAM (Ours)} &  & 69.47& 51.10& 72.14& 93.05& 73.97& 87.43& 69.62& 90.13& 75.86 \\
        \textbf{SEG-SAM2 (Ours)} &  & 68.33& 52.28& 72.43& 92.52& 73.67& 87.69& 69.83& 90.88& 75.95 \\
        \midrule
        Med2D* \citep{sammed2d} & \multirow{4}{*}{Box} & 70.33&	54.64&	81.71&	\textbf{94.67}&	72.19&	92.47&	88.26&	84.31&	79.82 \\
        MedSAM* \citep{medsam} &  & 72.41&	63.58&	78.56&	91.35&	66.44&	90.82&	86.35&	78.37&	78.49 \\
        Med-SA* \citep{medsa} &  & 72.12&	62.25&	80.94&	91.49&	71.86&	91.06&	86.74&	84.83&	80.16 \\
        \textbf{SEG-SAM (Ours)} &  & \textbf{79.45} & \textbf{64.99} & \textbf{83.67} & 94.51 & \textbf{75.53} & \textbf{92.70} & \textbf{88.39} & \textbf{90.44} & \textbf{83.46} \\
        \textbf{SEG-SAM2 (Ours)} &  & \textbf{78.36} & \textbf{65.14} & \textbf{83.72} & 94.02 & \textbf{75.29} & \textbf{92.98} & \textbf{88.63} & \textbf{90.57} & \textbf{83.59} \\
        \textbf{SEG-SAM3 (Ours)} &  & \textbf{79.50} & \textbf{65.35} & \textbf{84.04} & \textbf{94.92} & \textbf{76.60} & \textbf{93.50} & \textbf{89.26} & \textbf{91.58} & \textbf{84.34} \\
        \bottomrule
    \end{tabular}
\end{center}
\end{table*}

\subsubsection{Extensibility to SAM2 and SAM3}
\label{sec:Exten}
While instantiated on the original SAM, our proposed \textit{semantic-aware decoder} and \textit{text-to-vision semantic enhancement} are highly modular and can be adapted to SAM2 and SAM3 with minimal modifications.
Since the static-image variant of SAM2 exhibits architectural consistency with the original SAM, we directly integrate SAM2's image encoder into our framework. Combined with the proposed SAWD and T2VSE, this extension constructs SEG-SAM2.
Extending to SAM3 is slightly more complex since it introduces the \textit{text encoder} for text prompt generation and the \textit{detector} for mask prediction.
To adapt to this design, we first integrate SAWD as \textit{a parallel branch to SAM3's detector}. To ensure structural alignment, we adopt a 6-layer transformer identical to that of \textit{detector} while preserving the original design of the prediction head.
Then, prior to the \textit{detector} module, we embed T2VSE for independent text-to-vision interaction.
This process differs from SAM3's original text prompt-guided interaction in two main aspects: 1) we employ comprehensive medical descriptions instead of noun phrases, and 2) we utilize global-object text descriptions to enhance representation learning, rather than specific-object text prompt for caption-driven segmentation.
This difference also ensures compatibility with the SAM3 framework, so we can maintain both interactions during training but deliberately bypass SAM3's text prompts during inference to prevent semantic leakage. 
Unless otherwise specified, official default settings are used for remaining components to yield our SAM3 variant, denoted as SEG-SAM3, as shown in Fig. \ref{fig:figure_sam3}.

\subsection{Model Learning}
\label{sec:training}
We train SEG-SAM by simulating the interactive approach following \citep{sammed2d, sam}. Specifically, during training, we randomly select five foreground objects in each image to form a training batch, and for each object, we randomly sample either a foreground point or a bounding box (surrounding the object) with equal probability as the visual prompt. During inference, we randomly select one foreground object for segmentation, then generate a corresponding point or box prompt. 
In addition, SEG-SAM can be flexibly adapted for different inference tasks: for binary medical segmentation, we use $\mathrm{Dec_{sag}}$ to produce binary masks, while for semantic medical segmentation, we leverage the output of $\mathrm{Dec_{saw}}$ instead.
To ensure training stability, we additionally introduce two auxiliary losses: a \textit{cross-mask spatial alignment loss} to constrain consistency between two mask prediction branches $\mathrm{Dec_{sag}}$ and $\mathrm{Dec_{saw}}$, and an \textit{inter-token diversity loss} to encourage text summary tokens in T2VSE to capture distinct representational patterns.
In the following, we detail two loss functions and formulate the overall training objective of SEG-SAM.

\noindent \textbf{Cross-mask spatial alignment loss.}
In SEG-SAM, the two decoders learn complementary representations: $\mathrm{Dec_{saw}}$ for category semantic distinctions and $\mathrm{Dec_{sag}}$ for foreground-background separation.
Their predicted masks for the same prompted objects should align.
Therefore, we propose a \textit{cross-mask spatial alignment loss} to establish mask consistency of both predictions.
Given the respective mask outputs $M_s$ and $M_b$, we compute their intersection-over-union (IoU) and formulate the loss by minimizing $1-\mathrm{IoU}$ (theoretically, their IoU should be 1), as follows:
\begin{equation}
\label{Eq. 5}
\mathcal{L}^{cons} = 1 - \frac{|M_{s} \cap M_b|}{|M_{s} \cup M_b|}.
\end{equation}

\noindent \textbf{Inter-token diversity loss.}
To encourage the refined text summary tokens $t_{ts}'$ obtained from T2VSE to capture distinct representational patterns and thereby enhance model robustness, we introduce an \textit{inter-token diversity loss}.
Specifically, we compute the pairwise cosine similarity between non-identical tokens and penalize values that exceed a predefined margin $\mu$, as follows:
\begin{equation}
\label{Eq. tokenloss}
\mathcal{L}^{div} =
\operatorname*{mean}_{i\neq j}
\left[\max\left(0,\ \mathrm{CosSim}(t_{ts,i}',t_{ts,j}')-\mu\right)\right]^2,
\end{equation}
\noindent where $t_{ts,i}$ and $t_{ts,j}$ represent the $i$-th and $j$-th individual tokens in $t_{ts}'$, respectively.
The operator $\mathrm{mean}_{i \neq j}$ specifies averaging the penalty over all pairs of distinct tokens ($i \neq j$), and the squared hinge design imposes larger penalties on token pairs with higher similarity.

\begin{figure*}
\begin{center}
\includegraphics[width=0.95\linewidth]{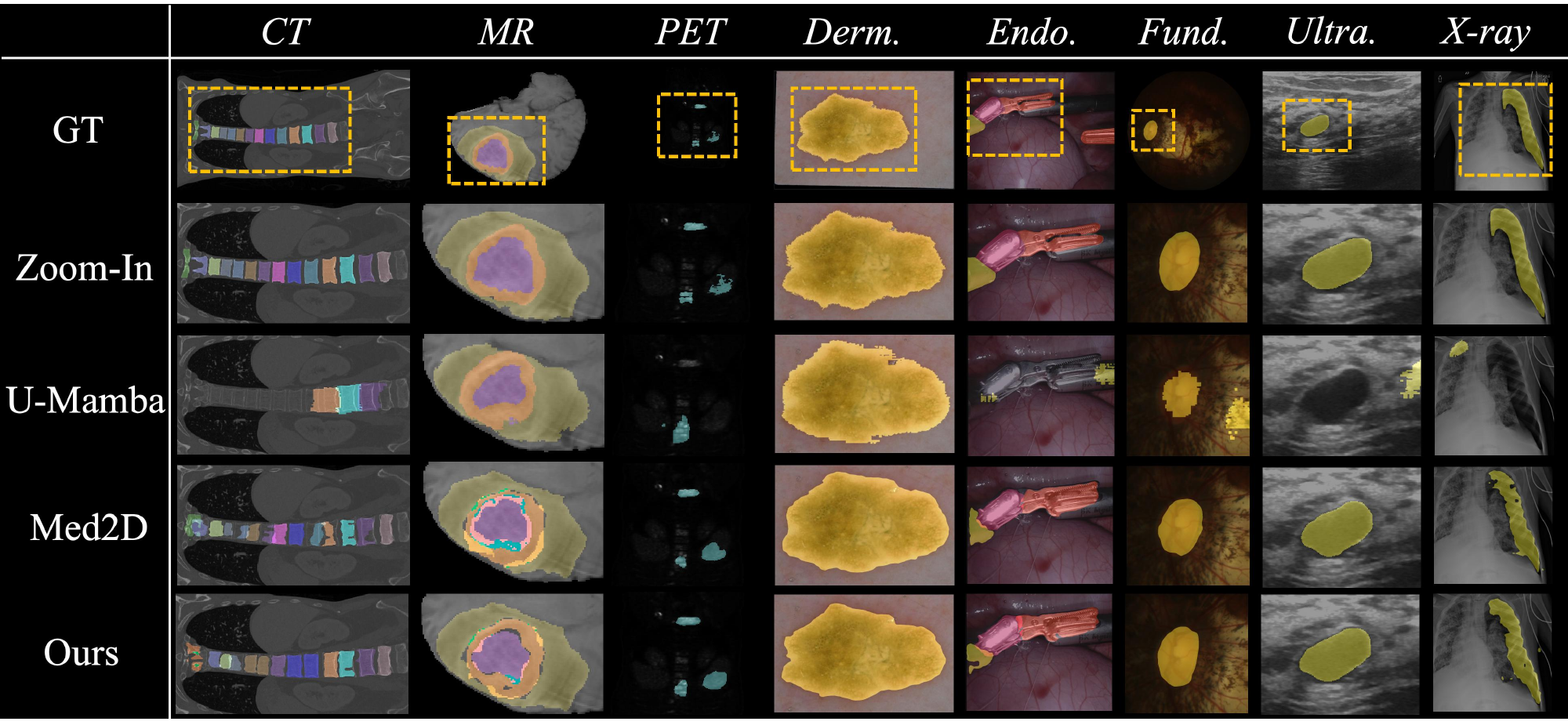}
\end{center}
\caption{Qualitative comparison between our method, U-Mamba \cite{umamba}, and Med2D \cite{sammed2d} across eight medical modalities. From top to bottom, we have the complete ground truth (GT), the zoom-in views of GT, and the zoom-in segmentation results obtained by U-Mamba, Med2D, and ours, respectively.
}
\label{fig:figure_vis}
\end{figure*}

\begin{figure}
\begin{center}
\includegraphics[width=1.0\linewidth]{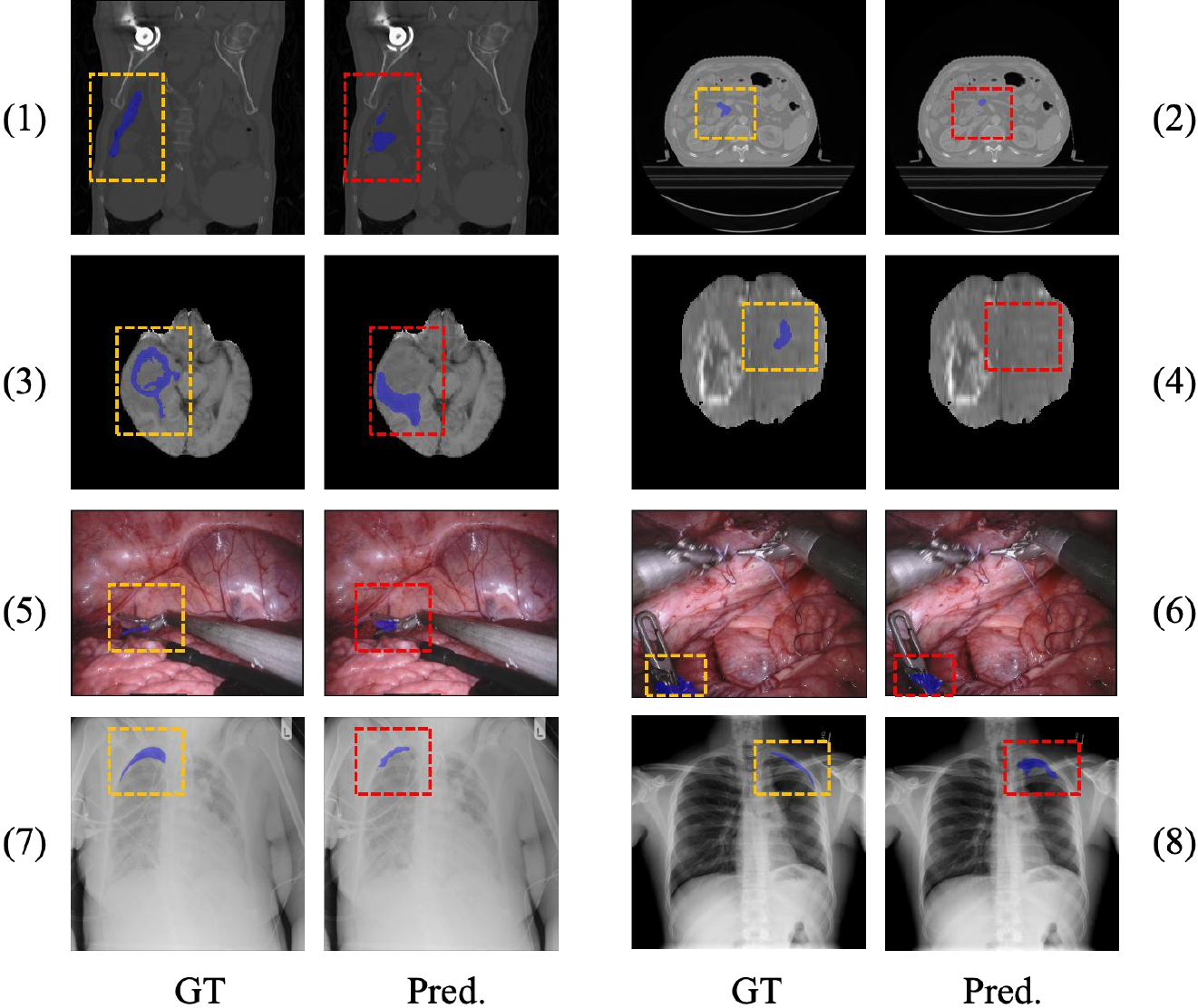}
\end{center}
\caption{
Failure case analysis of our method on the Med2D-16M dataset.
Yellow boxes indicate the ground truth, while red ones indicate the model's predictions. 
}
\label{fig:figure_badcase}
\end{figure}

\noindent \textbf{Overall loss.}
We employ the binary segmentation loss $\mathcal{L}^{bin}$ in $\mathrm{Dec_{sag}}$, which includes the focal loss $\mathcal{L}^{focal}$ and the Dice loss $\mathcal{L}^{dice}$ \citep{sam}. In $\mathrm{Dec_{saw}}$, we introduce semantic learning, consisting of both semantic mask prediction and semantic classification. For the former, we use the semantic segmentation loss $\mathcal{L}^{sem}$ \citep{mask2former}, which combines the cross-entropy loss $\mathcal{L}^{ce}$ and the segmentation Dice loss $\mathcal{L}^{dice}$. For the latter, we apply the binary cross-entropy loss $\mathcal{L}^{bce}$ for supervision.
The overall loss function of SEG-SAM can be written as:
\begin{equation}
\label{Eq. 5}
\mathcal{L}=\mathcal{L}^{bin} + \mathcal{L}^{sem} + \mathcal{L}^{cons} + \lambda_1 \mathcal{L}^{bce} + \lambda_2 \mathcal{L}^{div},
\end{equation}
\noindent where $\lambda_1$ and $\lambda_2$ are the loss weights of $\mathcal{L}^{bce}$ and $\mathcal{L}^{div}$.

\section{Experiments}
\subsection{Experimental setup}
\noindent \textbf{Datasets.}
Our experiments are conducted on the Med2D-16M dataset \citep{sammed2d, med2d16m}, which is recognized as the largest publicly available benchmark for medical image segmentation. It consists of eight modalities and approximately 200 categories, totaling 3.6M images and 16M annotated masks.
We split the dataset into an $80\%$ training set and a $20\%$ testing set.
After data preprocessing and cleaning, 191 categories are ultimately retained.
To evaluate the generalizability of our SEG-SAM, we conduct cross-dataset experiments on six external datasets.
The first group comprises three CT datasets with seen categories: KiTS23 \citep{kits}, BTCV \citep{btcv}, and AMOS \citep{amos}. The second group includes diverse modalities with unseen categories, such as the MR dataset CrossMoDA \citep{crossmoda}, the MR dataset HNTSMRG \citep{hntsmrg}, and the histopathology dataset EBHI \citep{ebhi}.

\noindent \textbf{Data pre-processing.}
For data pre-processing, we follow the same pre-processing approach in SAM-Med2D \citep{sammed2d}.
For 3D data, we slice them into 2D slices.
Specifically, modalities such as CT, MR, and PET primarily consist of 3D medical volumes, and each of these volumes often represents a complex scene containing multiple medical objects or anatomical structures. Each volume is sliced along the X, Y, and Z axes.
For 2D data, we simply ensure that all pixel values are within the range of $[0,255]$.
Additionally, we remove ground-truth masks whose area is less than 0.153\% (\textit{i.e.}, $100/(256 \times 256)$), as such masks are too small to extract \citep{sammed2d}.

\begin{table}
\begin{center}
    \caption{
    Comparison between our method, conventional methods, and state-of-the-art SAM-based methods on the KiTS, BTCV, and AMOS datasets. We use the bounding box as the visual prompt for SAM-based methods, and use the mDSC to evaluate the semantic segmentation performance. The "*" symbol indicates that we integrate our semantic-aware decoder with them.}
    \label{tab:cross_dataset_seen}
    \begin{tabular}{l |c c c}
        \toprule
        Method  & KiTS & BTCV & AMOS \\
        \midrule
        nnU-net &48.26&	24.33&	32.75 \\
        UNETR  &55.78&	23.02&	35.36 \\
        nnFormer  &52.17&	25.43&	38.94 \\
        U-Mamba  &54.51&	28.66&	40.19 \\
        \midrule
        Med2D*  & 72.61&	68.23&	66.78 \\
        MedSAM*  & 72.03&	66.74&	64.37 \\
        Med-SA*  & 73.16&	67.45&	65.34 \\
        \midrule
        \textbf{Ours} & \textbf{75.74} & \textbf{79.91} & \textbf{79.36} \\
        \bottomrule
    \end{tabular}
\end{center}
\end{table}

\begin{table}
\begin{center}
    \caption{
    Comparison between our method and state-of-the-art SAM-based methods on the HNTSMRG, CrossMoDA, and EBHI datasets. We use the bounding box as the visual prompt for SAM-based methods, and use the mDSC to evaluate the semantic segmentation performance. The "*" symbol indicates that we integrate our semantic-aware decoder with them.}
    \label{tab:cross_dataset_unseen}
    \begin{tabular}{l |c c c }
        \toprule
        Method  & HNTSMRG & CrossMoDA & EBHI \\
        \midrule
        Med2D*  & 47.83 & 64.45 & 71.29 \\
        MedSAM*  & 47.15 & 63.91 & 71.96 \\
        Med-SA*  & 48.51 & 64.02 & 71.43 \\
        \midrule
        \textbf{Ours} & \textbf{51.53} & \textbf{67.28} & \textbf{73.89} \\
        \bottomrule
    \end{tabular}
\end{center}
\end{table}

\noindent \textbf{Evaluation metrics.}
We adopt the widely used Dice Similarity Coefficient (DSC, also referred to as DICE), following previous works \citep{sammed2d, medsam} to evaluate the binary segmentation performance of our method. 
In addition, to comprehensively evaluate SEG-SAM's semantic segmentation, we further employ the widely used mean DICE (mDSC) \citep{deeplab, unet++} across different modalities for semantic medical segmentation.

\begin{figure}
\begin{center}
\includegraphics[width=0.95\linewidth]{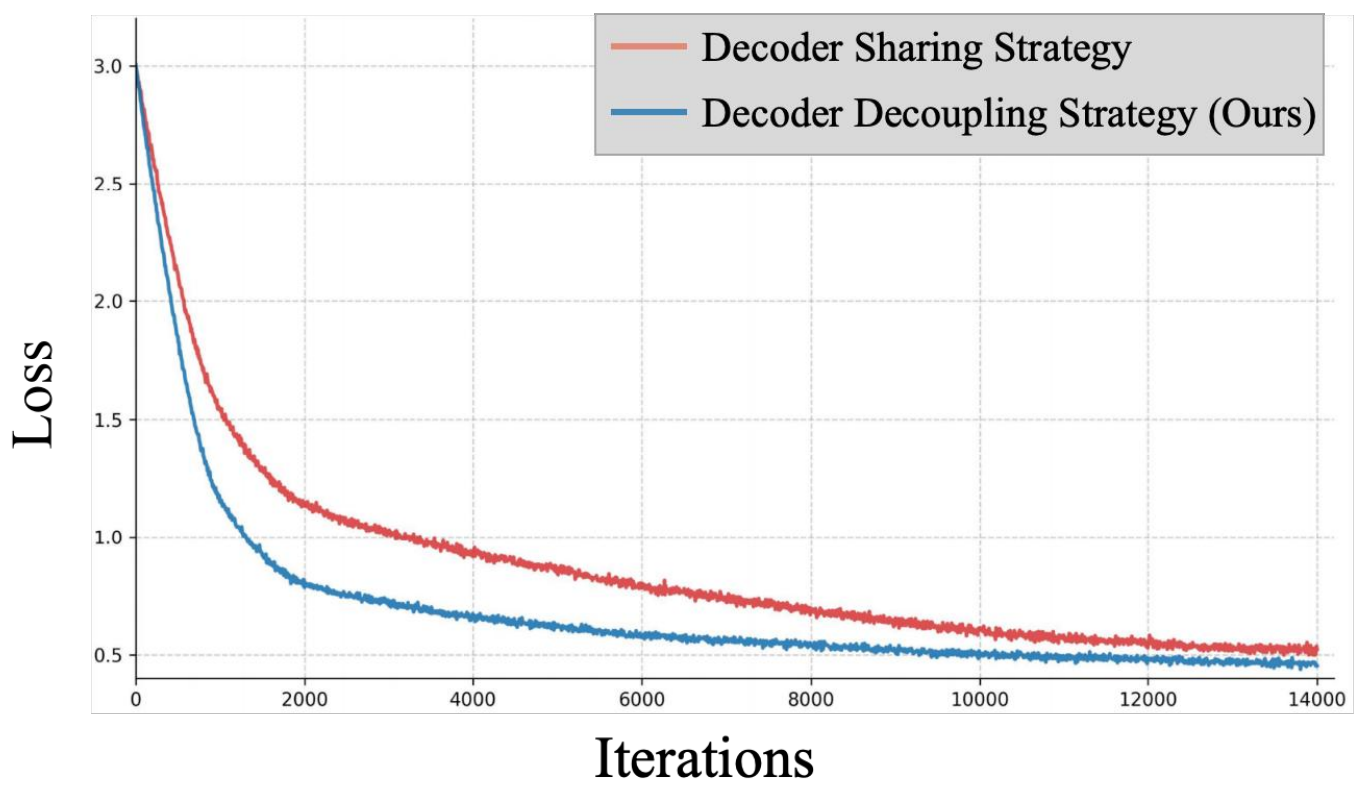}
\end{center}
\caption{Visualization of training loss curves for the decoder sharing and decoupling strategies.}
\label{fig:figure_loss}
\end{figure}

\begin{figure}
\begin{center}
\includegraphics[width=1.0\linewidth]{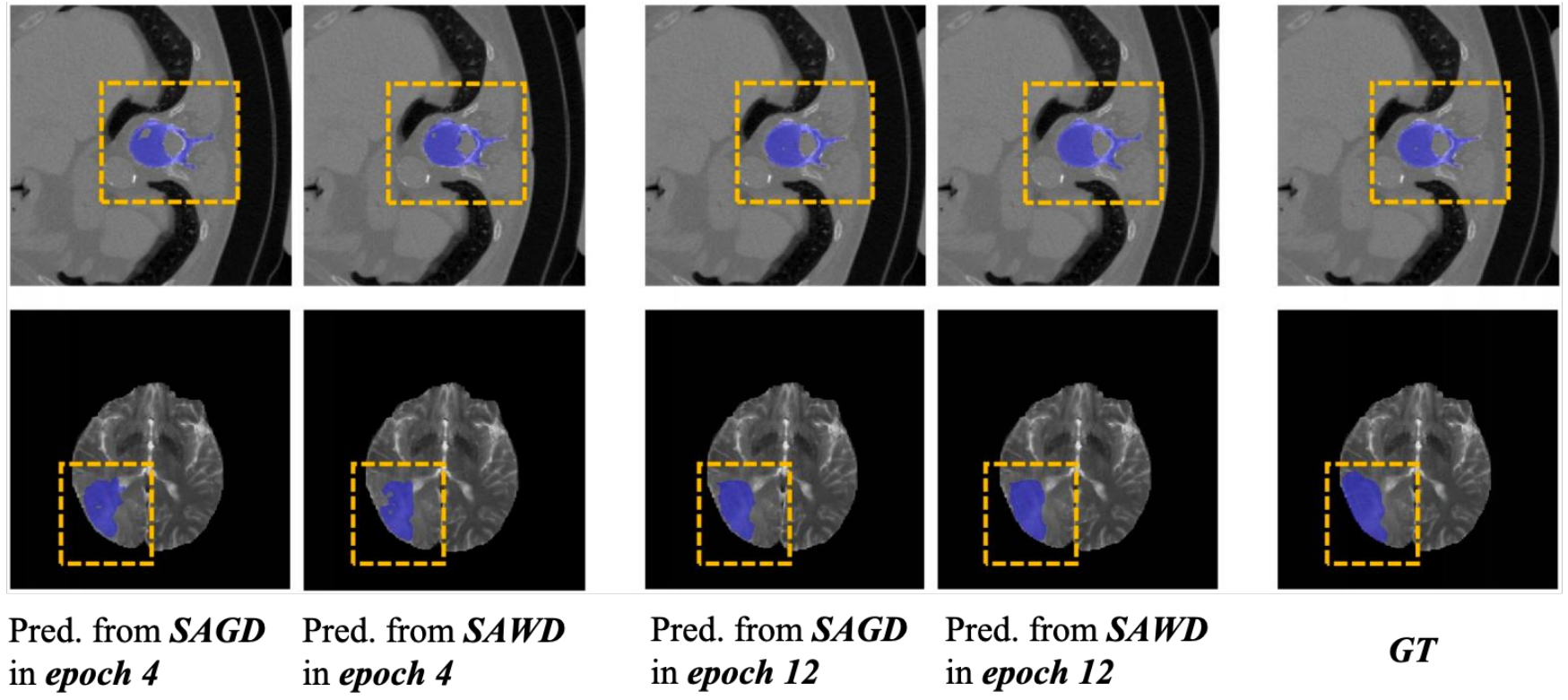}
\end{center}
\caption{Visualization of outputs from the semantic-agnostic decoder (SWGD) and the semantic-aware decoder (SAWD). As training propresses, the predictions of both decoders tend to converge.}
\label{fig:figure_maskalign}
\end{figure}

\noindent \textbf{Implementation details.}
In all experiments, we use 8 NVIDIA 4090 GPUs for training SEG-SAM and other methods, with each GPU processing 32 images per batch. 
All input images are resized to $256 \times 256$ pixels. 
We adopt the Adam optimizer \citep{adam} with an initial learning rate of $1e^{-4}$. 
The model is trained for 12 epochs, with the learning rate reduced by half at the 5th and 10th epochs.
For the image encoder, we use the pre-trained ViT-B/16 model from SAM \citep{sam} as the backbone and freeze its original parameters.
For the prompt encoder and mask decoder, we adopt the configurations from SAM and update their parameters during training.
For SAM3 and SEG-SAM3, we freeze the visual and text encoders and train the remaining components using the same training strategy.
Empirically, we set the loss balancing weights to $\lambda_1 = 0.5$ and $\lambda_2 = 0.01$, and the margin threshold to $\mu = 0.9$.
A relatively small weight is assigned to $\lambda_2$ because $\mathcal{L}^{div}$ acts as an auxiliary regularizer; an overly large weight could overemphasize token diversity, thereby interfering with the primary optimization of text summary tokens in T2VSE.
Furthermore, the parameter for the TopK operation ($K=16$) and the number of text summary tokens ($M=4$) are determined based on subsequent ablation studies.

\subsection{Unified medical segmentation}
\noindent \textbf{Unified binary medical segmentation.}
We first evaluate the binary medical segmentation performance of our method on the Med2D-16M dataset. In Table \ref{tab:bin_seg}, we compare SEG-SAM with vanilla SAM \citep{sam} without fine-tuning and its decoder fine-tuned version (FT-SAM).
For SAM2 \citep{sam2}, we focus on the decoder fine-tuned SAM2 and our proposed variant (SEG-SAM2). 
We also compare SEG-SAM with three state-of-the-art SAM-based methods, including SAM-Med2D \citep{sammed2d}, MedSAM \citep{medsam}, and Med-SA \citep{medsa}.
All these methods are implemented following their publicly available codebase and apply the same interactive training strategy through point or bounding box prompts.
In addition, we extend our method to the SAM3 \citep{sam3} framework, comparing its fine-tuned baseline (FT-SAM3) with our proposed variant (SEG-SAM3). Following the official codebase, only bounding boxes are employed as visual prompts.
We employ the DSC metric to evaluate the binary medical segmentation performance across 8 modalities, and additionally report the average DSC over all modalities.
The results demonstrate that SEG-SAM consistently outperforms other SAM-based methods across all modalities. Notably, it achieves significant improvement over the second-best approach regardless of whether point-based (73.36 vs. 68.48) or box-based (81.85 vs. 78.64) prompts are used.
Specifically, our method achieves substantial gains across different modalities, including CT {2.7\%}, MR {4.0\%}, PET {2.4\%}, and X-ray {5.7\%}.
Even in data-scarce modalities such as Dermoscopy and Fundus, SEG-SAM maintains a performance lead of $1.0\%$.
Furthermore, SEG-SAM2 yields only slightly better overall performance than SEG-SAM.
Benefiting from its strong architecture and extensive pre-training, FT-SAM3 achieves highly competitive results, second only to SEG-SAM.
Furthermore, our SAM3-based extended variant, SEG-SAM3, attains the highest overall performance of 82.71, surpassing the SEG-SAM (81.85) by 1.0\%. This clearly demonstrates the effectiveness and scalability of our proposed method.

\begin{table}
\begin{center}
    \footnotesize
    \caption{
    Ablation study on data scale in the training set. We conduct experiments on the Med2D dataset and evaluate semantic segmentation using the mDSC metric.}
    \label{tab:table_scale}
    \begin{tabular}{l |c c c c c}
        \toprule
         Data scaling & 1/16 & 1/8 & 1/4 & 1/2 & All \\
        \midrule
        
        Point Prompt & 67.26 & 71.07 & 73.42 & 74.91 & 75.86 \\
        Box Prompt & 76.24 & 78.32 & 80.97 & 82.04 & 83.46 \\
        \bottomrule
    \end{tabular}
\end{center}
\end{table}

\begin{table}
\begin{center}
    \caption{
    Ablation study of the decoder decoupling and decoder sharing strategies in our SEG-SAM. We conduct experiments on the Med2D-16M dataset and take the vanilla SAM as Baseline for binary segmentation, while introducing a classification token in SAM's decoder as Baseline$_{sem}$ for semantic segmentation. We use DSC and mDSC to evaluate binary and semantic segmentation performance, respectively.}
    \label{tab:table_abde}
    \begin{tabular}{l |c c }
        \toprule
        Decoder Architecture  & DSC & mDSC  \\
        \midrule
         $\mathrm{Baseline}$  & 78.44 & -\\
         $\mathrm{Baseline_{sem}}$ & 76.46 & 77.39  \\
        Decoder sharing & 79.28 & 80.33 \\
       {Decoder decoupling (Ours)}  & 81.85 & 83.46 \\
        \bottomrule
    \end{tabular}
\end{center}
\end{table}

\begin{figure}
\begin{center}
\includegraphics[width=1.0\linewidth]{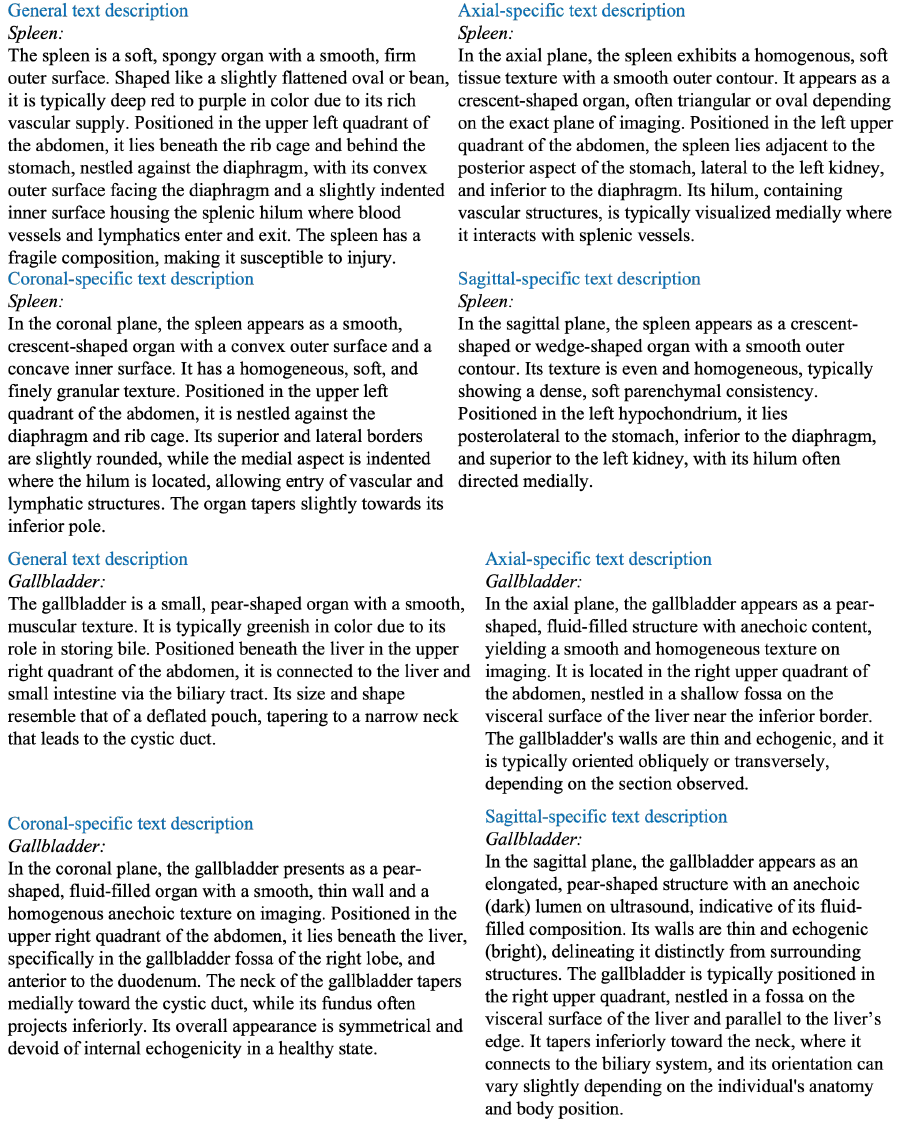}
\end{center}
\caption{Examples of general text descriptions and view-specific text descriptions given the Spleen and Gallbladder categories generated by an LLM (ChatGPT).}
\label{fig:figure_textdesciption}
\end{figure}

\noindent \textbf{Unified semantic medical segmentation.}
We further evaluate the semantic segmentation performance of our three models on the Med2D-16M dataset. In Table \ref{tab:sem_seg}, we compare SEG-SAM with SEG-SAM2 and SEG-SAM3 with conventional medical segmentation methods, including nnU-net \citep{nnunet}, UNETR \citep{unetr}, nnFormer \citep{nnformer}, and U-Mamba \citep{umamba}. 
For a fair comparison, we train these methods using the same setup as our method on the Med2D-16M dataset.
Furthermore, we integrate the proposed semantic-aware decoder-based learning strategy into the SAM-based methods, including Med2D \citep{sammed2d}, MedSAM \citep{medsam}, and Med-SA \citep{medsa}, thereby extending them for unified semantic medical segmentation.
We employ the mDSC metric to evaluate the performance of all methods across 8 modalities and report the average mDSC.
The results demonstrate that our method significantly outperforms both task-specific methods and other SAM-based methods in terms of average metrics across all modalities.
Under box-based prompts, SEG-SAM achieves substantial performance improvements over the second-best method, particularly in CT {9.7\%}, MR {2.2\%}, and X-ray {6.6\%}.
Furthermore, SAM-based methods integrated with our semantic learning strategy (marked with * in Table \ref{tab:sem_seg}) also demonstrate consistent performance gains. 
These enhanced models achieve mDSC scores of 79.82, 78.49, and 80.16, respectively, significantly outperforming the best traditional method 68.98.
Lastly, consistent with the binary segmentation results, SEG-SAM2 achieves slightly better performance than SEG-SAM and SEG-SAM3 maintains its advantage in semantic task, achieving the highest overall performance of 84.34 across different modalities.

\begin{table*}
\centering
\caption{
Ablation study of key components in SEG-SAM on the Med2D-16M dataset. 
We add each of our components to the Baseline and use DSC and mDSC to evaluate the performance improvement for binary and semantic segmentation, respectively.}
\label{tab:sawd}
\begin{tabular}{ccccc|cc}
\toprule

\multirow{2}{*}{Method} & \multicolumn{4}{c}{Components} & \multicolumn{2}{|c}{Metrics} \\ 
& $\mathrm{Dec_{saw\_so}}$ & $\mathrm{Dec_{saw\_co}}$ & T2VSE & $\mathcal{L}^{cons}$ & DSC & mDSC \\

\midrule
$\mathrm{Baseline}$ &  &  &  &    & 78.44 & - \\
$\mathrm{Baseline_{sem}}$ &  &    &  &  & 76.46 & 77.39 \\
\midrule
1) & \checkmark &  &   &  & 80.12 & 81.39 \\
2) & \checkmark & \checkmark &    &  & 80.85 & 82.03 \\
3) & \checkmark & \checkmark &   \checkmark & & 81.40 & 83.19 \\
\midrule
Ours & \checkmark & \checkmark &  \checkmark  & \checkmark & 81.85 & 83.46 \\ 

\bottomrule
\end{tabular}
\end{table*}

\noindent \textbf{Cross-dataset experiments.}
As a foundation model, SAM inherently possesses zero-shot segmentation capability. To verify whether our method retains this capability, we conduct comprehensive cross-dataset evaluations across diverse imaging modalities.
First, we introduce three new datasets for evaluation, including the KiTS23 \citep{kits}, BTCV \citep{btcv}, and AMOS \citep{amos} datasets, which are strictly excluded from the unified Med2D-16M dataset, yet their object categories are covered within its label set. 
We use the mDSC metric to evaluate the semantic medical segmentation performance.
In Table \ref{tab:cross_dataset_seen}, we test the trained model from Med2D-16M on these datasets.
All comparable methods are trained in the same way.
Experimental results indicate that SAM-based methods exhibit stronger cross-dataset generalization capabilities compared to task-specific methods, and our method achieves the best performance across three CT datasets.
Benefiting from the enhanced visual representations via our decoupled semantic learning strategy, SEG-SAM achieves substantial improvements of {3.5\%}, {17.1\%}, and {18.8\%} on these datasets, respectively.

Furthermore, we introduce three additional datasets with object categories not included in Med2D-16M: HNTSMRG \citep{hntsmrg}, CrossMoDA \citep{crossmoda}, and EBHI \citep{ebhi}.
Notably, our method can generalize to object categories beyond the training set without requiring additional training.
We utilize the identical LLM to generate and extract their corresponding text embeddings. 
The obtained embeddings are then fed into SAWD for classification by calculating the cosine similarity with the visual features.
Based on this, other SAM-based methods integrated with SAWD can achieve the same predictive capability through this mechanism.
We use the mDSC metric to evaluate semantic medical segmentation performance.
In Table \ref{tab:cross_dataset_unseen}, we test the trained models from Med2D-16M on these datasets.
All comparable methods are trained in the same way.
Experimental results demonstrate that our method consistently outperforms other SAM-based methods.

\begin{table}
\begin{center}
    \footnotesize
    \caption{
    Ablation study on the design of semantic-aware decoder. We conduct experiments on the Med2D-16M dataset and use DSC, mDSC, Hours (h), and FPS to measure binary segmentation, semantic segmentation, total training time, and average inference speed, respectively.}
    \label{tab:table_cls_seg}
    \begin{tabular}{l |c c c c }
        \toprule
         Decoder & DSC & mDSC & Train Time & Infer. Sp. \\
        \midrule
        
        $\mathrm{Dec_{saw\_segall}}$ & 81.69 & 83.14 & 210 h & 21.7 FPS \\
        $\mathrm{Dec_{saw}}$ (Ours) & 81.85 & 83.46 & 120 h & 24.2 FPS\\
        \bottomrule
    \end{tabular}
\end{center}
\end{table}

\begin{table}
\begin{center}
    \caption{
    Ablation study on the plug-and-play capability of our semantic-aware decoder. We integrate it into other SAM-based methods and conduct experiments on the Med2D-16M, KiTS, and BTCV datasets. We use DSC to evaluate the binary segmentation performance. The "*" symbol indicates these methods are integrated with our semantic-aware decoder.}
    \label{tab:table4}
    \begin{tabular}{l|c c c}
        \toprule
        Method & Med2d & KiTS & BTCV \\
        \midrule
        SAM-Med2D & 78.44&	86.91&	83.30 \\
        SAM-Med2D* & \textbf{80.34} & \textbf{87.41} & \textbf{84.77} \\
        \midrule
        MedSAM & 78.64&	85.89&	83.72 \\
        MedSAM* & \textbf{80.61} & \textbf{86.64} & \textbf{84.37} \\
        \midrule
        Med-SA & 78.53&	86.18&	82.83 \\
        Med-SA* & \textbf{79.91} & \textbf{86.72} & \textbf{84.03} \\
        \bottomrule
    \end{tabular}
\end{center}
\end{table}

\noindent \textbf{Data scaling.}
To further investigate the impact of data scale on our model performance, we conduct a data scalability experiment.
We create five subsets via stratified sampling with ratios of $1/16$, $1/8$, $1/4$, $1/2$, and $100\%$ of the original SAM-Med2D dataset, ensuring that all categories are included in each subset. 
Results in Table~\ref{tab:table_scale} demonstrate that as the dataset scale increases, model performance steadily improves. Notice when the data scale goes beyond $1/4$, the performance increase slows down a bit.

\noindent \textbf{Visualization results.}
In Fig. \ref{fig:figure_vis}, we visualize the semantic segmentation results of our method on the Med2D-16M dataset, comparing it with Med2D \citep{sammed2d} and U-Mamba \citep{umamba}.
While our SEG-SAM demonstrates impressive performance on unified medical segmentation tasks across diverse modalities and categories, our visualization analysis reveals limitations in two challenging scenarios: segmenting structures with ambiguous anatomical boundaries and extremely tiny objects. As illustrated by the failure cases in Fig.~\ref{fig:figure_badcase}, the model's performance degrades in the presence of low contrast between adjacent tissues or overlapping anatomical structures (\textit{e.g.}, cases (1), (6), and (8)). Furthermore, our SEG-SAM exhibits difficulty in accurately segmenting tiny objects, as exemplified in cases (2) and (5), where fine-grained details are occasionally missed.

\subsection{Ablation study}
\label{sec:ablation study}
Unless otherwise stated, we conduct the ablation study on the Med2D-16M dataset to evaluate different components in SEG-SAM by using the bounding boxes as the visual prompt. We use DSC and mDSC to evaluate binary and semantic segmentation performance, respectively.
Specifically, a vanilla SAM-Med2D serves as our Baseline only for binary segmentation.
Additionally, we introduce a learnable token in SAM-Med2D's mask decoder, transforming it into a semantic decoder, which serves as another Baseline$_\text{sem}$ only for semantic segmentation.

\noindent \textbf{Decoder decoupling strategy.}   
To verify the effectiveness of our designed decoder decoupling strategy, we present a variant by modifying the original mask decoder to simultaneously handle both binary and semantic segmentation tasks.
This variant, denoted by \textit{decoder sharing} in Table~\ref{tab:table_abde},
shares most of the decoder layers, with two independent prediction heads (each consisting of several convolutional layers and MLPs), and employs two types of learnable tokens to predict binary and semantic masks, respectively.
The results indicate that simply using the \textit{decoder sharing} strategy to learn both tasks is insufficient, yielding only marginal improvement of 0.84 in DSC from $\mathrm{Baseline}$ and 2.94 in mDSC from $\mathrm{Baseline_{sem}}$. In contrast, our SEG-SAM leveraging the \textit{decoder decoupling} strategy results in substantial performance improvement compared to \textit{decoder sharing} (3.41 vs. 0.84 in DSC and 6.07 vs. 2.94 in mDSC), which validates the motivation behind our design.
Furthermore, we illustrate both loss curves for the decoder sharing and decoder decoupling strategies in Fig.~\ref{fig:figure_loss}, where one can clearly see that the former converges more slower and is inferior to the latter.

\begin{table}
\begin{center}
    \caption{Comparison of the effectiveness of injecting $t_{ts}$ into $\mathrm{Dec_{sag}}$ and $\mathrm{Dec_{saw}}$, the number of $t_{ts}$ (M), and the TopK parameter (K). We conduct experiments on the Med2D-16M dataset and use DSC and mDSC to evaluate binary and semantic segmentation performance, respectively.}
    \label{tab:table_tts}
    \begin{tabular}{l |c c }
        \toprule
        Method  & DSC & mDSC  \\
        \midrule
         w/o injecting $t_{ts}$ & 81.20 & 82.13 \\
         Injecting $t_{ts}$ into $\mathrm{Dec_{saw}}$ & 81.22 & 82.17 \\
         Injecting $t_{ts}$ into $\mathrm{Dec_{sag}}$ (Ours) & 81.85 & 83.46 \\
         Injecting $t_{ts}$ into $\mathrm{Dec_{sag}}$ and $\mathrm{Dec_{saw}}$ & 81.85 & 83.48 \\
         \midrule
         \multicolumn{3}{l}{\textbf{Injecting $t_{ts}$ into $\mathrm{Dec_{sag}}$}} \\
         \quad $M=1$ & 81.74 & 83.33 \\
         \quad $M=2$ & 81.81 & 83.41 \\
         \quad $M=4$ (Ours) & 81.85 & 83.46 \\
         \quad $M=8$ & 81.71 & 83.37 \\
         \multicolumn{3}{l}{\textbf{Injecting $t_{ts}$ into $\mathrm{Dec_{sag}}$}} \\
         \quad $K=8$ & 81.77 & 83.39 \\
         \quad $K=16$ (Ours) & 81.85 & 83.46 \\
         \quad $K=32$ & 81.80 & 83.43 \\
        \bottomrule
    \end{tabular}
\end{center}
\end{table}

\begin{table}
\begin{center}
    \caption{
    Comparison of three text templates in the text-to-vision semantic enhancement scheme, including the Name-original based on the ChatGPT, Gemini, and Claude, respectively, and Name-revised (Ours) strategies. We conduct experiments on the Med2D-16M dataset and use DSC and mDSC to evaluate binary and semantic segmentation performance, respectively.}
    \label{tab:table_t2vse}
    \begin{tabular}{l |c c }
        \toprule
        Text Template  & DSC & mDSC  \\
        \midrule
         Name-original-ChatGPT & 81.83 & 83.45 \\
         Name-original-Gemini & 81.83 & 83.44 \\
         Name-original-Claude & 81.82 & 83.43 \\
         Name-revised (Ours) & 81.85 & 83.46 \\
        \bottomrule
    \end{tabular}
\end{center}
\end{table}

\noindent \textbf{Semantic-aware decoder.}
\label{abla_saw}
To further evaluate $\mathrm{Dec_{saw}}$, starting from the Baseline in Table \ref{tab:sawd}, we gradually build $\mathrm{Dec_{saw}}$ by first introducing  the SO-Token into the decoder for semantic segmentation of the prompted object, denoted as $\mathrm{Dec_{saw\_so}}$; then we add CO-Tokens into the decoder for semantic classification of all objects in the image, denoted by $\mathrm{Dec_{saw\_co}}$. 
Results indicate that the SO-Token in $\mathrm{Dec_{saw}}$ brings semantic prediction capability into SAM and also improves its binary segmentation performance; adding the CO-Tokens further enhances the model performance. Overall, our $\mathrm{Dec_{saw}}$ leads to an increase in DSC by 2.41 compared to Baseline and in mDSC by 4.64 compared to $\mathrm{Baseline_{sem}}$.

Furthermore, to justify why we do not perform the semantic segmentation task on all objects in the image, we introduce a variant by doing this, denoted as $\mathrm{Dec_{saw\_segall}}$.
In this variant, CO-Tokens are employed to perform semantic segmentation for other non-prompted objects in the image.
Since both represent pixel-level tasks, we concatenate the CO-Tokens with the SO-Token for joint interaction with the visual features, but use separate prediction heads.
Then, based on the CO-Token predictions, we apply a mask pooling operation to extract object prototypes, which are then utilized to enhance the semantic segmentation of the prompted object.
Apart from the segmentation performance, we also measure the training time and average inference speed in Table \ref{tab:table_cls_seg}. Compared with $\mathrm{Dec_{saw\_segall}}$, our proposed $\mathrm{Dec_{saw}}$ reduces the training time by approximately $1.8\times$ and increases the average inference speed by $2.5$ FPS, while retaining slightly better segmentation performance.
The experimental results validate our motivation of separating the processing of the prompted object from the rest of objects in the image. 

Lastly, to validate the plug-and-play capability of the proposed semantic-aware decoder in enhancing binary segmentation, we integrate it into several SAM-based methods and evaluate the binary segmentation performance on three datasets with the DSC metric, including the Med2D-16M, KiTS, and BTCV datasets.
As shown in Table \ref{tab:table4}, the results demonstrate that SAWD effectively improves binary segmentation performance of the SAM-based frameworks, further verifying its scalability.

\noindent \textbf{Text-to-vision semantic enhancement.}
\label{ablation:t2vse}
To validate the effectiveness of text-to-vision semantic enhancement, we ablate it in Table \ref{tab:sawd} and 
results show that T2VSE improves both binary and semantic segmentation performance, leading to an improvement of 0.55 in DSC and 1.16 in mDSC.
Additionally, in T2VSE, we incorporate the text summary token $t_{ts}$ to capture medical knowledge and integrate it into the subsequent decoding process.
Due to the lack of class semantics, our $\mathrm{Dec_{sag}}$ outputs only a binary foreground-background mask.
$t_{ts}$ fills this gap by injecting object semantics based on the category text embeddings.
In contrast, in $\mathrm{Dec_{saw}}$, semantic cues are explicitly available from ground-truth labels, so the benefit of adding $t_{ts}$ is trivial.
To further evaluate this point, we conduct an ablation study to compare the segmentation performance of injecting $t_{ts}$ into $\mathrm{Dec_{sag}}$ and $\mathrm{Dec_{saw}}$.
As shown in Table~\ref{tab:table_tts}, injecting $t_{ts}$ into $\mathrm{Dec_{sag}}$ clearly helps while injecting it into $\mathrm{Dec_{saw}}$ leads to insignificant improvement.
Furthermore, to determine the optimal configuration, we evaluate the impact of varying the number of text summary tokens $t_{ts}$. As reported in Table \ref{tab:table_tts}, setting $M=4$ achieves the best performance.
We further investigate the influence of parameter $K$ in the TopK operation.
The results in Table \ref{tab:table_tts} indicate that setting $K=16$ yields the optimal segmentation accuracy.
Overall, the framework demonstrates low sensitivity to different values of $M$ and $K$, suggesting its robustness.

Notably, we generate text descriptions per category instead of per image. We replace [class name] in the text template (see Sec.~\ref{sec:Text}) with each predefined class name as input to ChatGPT, obtaining the text descriptions of all categories. This reduces randomness and uncertainty. Nevertheless, to further investigate the variability of generated descriptions across LLMs, we also use Claude and Gemini for category description generation, which are referred to as Name-original-ChatGPT, Name-original-Claude, and Name-original-Gemini, respectively. As shown in Table \ref{tab:table_t2vse}, the experimental results show that all methods produce similar results, demonstrating low variability of descriptions at the category level.
In addition, we have engaged medical experts to manually revise the text descriptions generated by ChatGPT, which are denoted as Name-revised.
The expert revisions are very trivial and offer very slight benefit.
Nevertheless, we prioritize the expert-revised versions as the default configuration to guarantee medical accuracy.

\begin{table}
\begin{center}
    \footnotesize
    \caption{
    Comparison general and view-specific text descriptions. We conduct experiments on the Med2D-16M testing set and use DSC to evaluate binary segmentation performance. "CA" denotes the cross-attention mechanism in T2VSE.}
    \label{tab:table_textCA}
    \begin{tabular}{l |c c c}
        \toprule
        Method  & X-axis & Y-axis & Z-axis  \\
        \midrule
         SEG-SAM w/o text & 75.52 & 74.25 & 74.19 \\
         SEG-SAM w/ text, w/o CA & 75.65 & 74.48 & 74.33 \\
         SEG-SAM w/ vs-text, w/o CA  & 75.72 & 74.59 & 74.41 \\
         SEG-SAM w/ text, w/ CA (def) & 76.03 & 74.82 & 74.69 \\
         SEG-SAM w/ vs-text, w/ CA & 76.03 & 74.84 & 74.71\\

        \bottomrule
    \end{tabular}
\end{center}
\end{table}

\begin{table}
\begin{center}
    \caption{
    Comparison across different 2D views. We conduct inference experiments on the Med2D-16M test set and use mDSC to evaluate semantic segmentation performance. "ax" denotes that text descriptions based on different slicing axes are used.}
    \label{tab:table_axes}
    \begin{tabular}{l |c c c}
        \toprule
        Method  & X-axis & Y-axis & Z-axis  \\
        \midrule
         CT modality & 80.73 & 78.24 & 77.85 \\
         MR modality& 65.79 & 65.26 & 64.80 \\
         PET modality& 84.29 & 83.14 & 83.48 \\
        \midrule
         CT modality w/ ax & 81.76 & 78.22 & 77.86 \\
         MR modality w/ ax & 65.80 & 65.23 & 64.81 \\
         PET modality w/ ax& 84.28 & 83.16 & 83.49 \\

        \bottomrule
    \end{tabular}
\end{center}
\end{table}

In our setup, a portion of the 2D slices are extracted from 3D medical volumes. Any 2D slice obtained from a 3D volume is basically a cross-sectional view of the 3D object. Although visual features in a 2D slice may not completely align with the generated text description, the slice represents a specifically localized manifestation of the 3D object. Hence, the text descriptions can still be relevant and useful to 2D slices.
Crucially, we acknowledge that using general text descriptions for view-specific 2D slices may introduce slight noise and misalignment. Nevertheless, the text-vision cross-attention (CA) mechanism incorporated in T2VSE's second stage can sufficiently mitigate this issue (see Sec.~\ref{sec:Text}). The text embeddings are selectively aggregated and integrated with visual information, where the potentially irrelevant or misaligned text knowledge is effectively suppressed in this process. 
To evaluate the robustness of CA and investigate whether view-specific text descriptions for 2D slices are necessary, we conduct an additional experiment on the Med2D-16M dataset.
Specifically, to generate view-specific text descriptions for images from different views, we instead employ three text prompts corresponding to the axial ($Z$-axis), coronal ($Y$-axis), and sagittal ($X$-axis) views: 
\textit{Please summarize the visual and physical characteristics of
[class name] in the [view name] in a short, fluent paragraph. Focus on its most salient properties that best define this category, such as texture, shape, and position.}
We split test images into three subsets according to the slicing axes (X, Y, and Z axes). We report the binary segmentation performance in five settings and retain the first stage interaction: (i) training without text descriptions (w/o text); (ii) training with general text descriptions but without CA (w/ text, w/o CA); (iii) training with view-specific text descriptions but without CA (w/ vs-text, w/o CA); (iv) training with general text descriptions and CA (default, w/ text, w/ CA); and (v) training with view-specific text descriptions and CA (w/ vs-text, w/ CA). Specifically, in (ii) and (iii), we remove the cross-attention mechanism but instead perform an element-wise addition between the text embeddings and resized visual features for subsequent processing. 
As shown in Table \ref{tab:table_textCA}, all variants from (ii) to (v) yield improvements over the baseline (i). 
This verifies that tissues and organs can sometimes share high inter-class visual similarities in medical images, while adding text descriptions for providing category-specific semantic guidance can help the model discriminate between these visually similar object classes. Moreover, comparing the results of (ii) and (iv) with those of (iii) and (v), we observe that the text-vision cross-attention is beneficial for suppressing irrelevant information from the text while further improving the performance; because of using it, leveraging view-specific text descriptions (v) and general text descriptions (iv) performs very closely. On the other hand, without the cross-attention, leveraging view-specific text descriptions (iii) performs slightly better than leveraging general descriptions (ii). We show some examples of the two types of descriptions in Fig.~\ref{fig:figure_textdesciption}, they do not present significant differences, as noted, using the general category text descriptions generated from an LLM is sufficient to benefit 2D slices.
Introducing detailed descriptions over different axes may potentially lead to overfitting, where the model relies heavily on view-specific details, while ignoring the overall anatomical structure.

\noindent \textbf{Cross-mask spatial alignment.}
To evaluate the impact of the cross-mask spatial alignment, we remove the consistency loss $\mathcal{L}^{cons}$ for model training.
We observe that this loss clearly benefits the model for both binary and semantic segmentation performance. As shown in Table \ref{tab:sawd}, removing this loss leads to a decrease of 0.45 in DSC and 0.27 in mDSC.
Furthermore, we provide visual examples in Fig.~\ref{fig:figure_maskalign} demonstrating how the alignment between the predictions, \textit{i.e.}, $M_{b}$ from $\mathrm{Dec_{sag}}$ and $M_{s}$ from $\mathrm{Dec_{saw}}$, improves throughout training due to the incorporation of this loss. These examples show that as the training progresses, the discrepancies between the binary and semantic masks are gradually reduced, leading to notably better alignment and consistency between the two. In addition, both masks proceed towards the ground truth.

While our motivation for introducing the cross-mask spatial alignment loss is to encourage better alignment between the two predictions and penalize their deviations, we also acknowledge that some inconsistencies may persist.
This is because the binary mask $M_{b}$ and the semantic mask $M_{s}$ are optimized through different pathways: $M_{b}$ aims to distinguish foreground regions from background, whereas $M_{s}$ focuses on discriminating specific semantic categories. It is reasonable to allow some degree of divergence between their predictions. This loss can reduce this divergence. However, we also do not put a big weight on this loss as otherwise it may excessively constrain the model and impede its generalizability.

\noindent \textbf{Multi-view data inference.}
To validate the robustness of our method across different 2D views, we conduct an additional experiment on the Med2D-16M testing set. Specifically, 2D slices from 3D volumes (CT, MR, and PET) are divided into three subsets based on their slicing axes: $X$, $Y$, and $Z$. We compare the semantic segmentation performance of our method on the 2D slices from each axis. The experimental results in Table~\ref{tab:table_axes} demonstrate that our method delivers comparable performance across slices from all three views, highlighting its robustness.
Furthermore, we conduct an inference experiment based on the text descriptions with different slicing axes (discussed in Sec.~\ref{ablation:t2vse}), with results denoted by the suffix ``w/ ax" in Table~\ref{tab:table_axes}. 
All other experimental settings remain consistent.
The results indicate that while the new descriptions lead to performance gains in some cases and declines in others, the overall improvement is insignificant.

\section{Conclusion}
In this work, we introduce SEmantic-Guided SAM (SEG-SAM), a unified model designed to enhance medical image segmentation performance by incorporating semantic knowledge into the SAM framework.
We first develop a semantic-aware decoder (SAWD) that complements SAM’s original semantic-agnostic decoder, enabling the model to handle both binary and semantic segmentation tasks within medical images. 
By integrating a text-to-vision semantic enhancement (T2VSE) scheme, SEG-SAM leverages detailed category-specific knowledge from a large language model to improve binary segmentation performance.
Additionally, we propose two constraint losses to ensure optimization stability, including a cross-mask spatial alignment strategy to promote consistency between the outputs of the two decoders and an inter-token diversity loss to encourage the text summary tokens to capture differential semantic patterns.
Notably, both SAWD and T2VSE are highly modular, allowing them to be flexibly embedded into advanced architectures such as SAM2 and SAM3 to enhance their medical segmentation performance.
Extensive experiments on multiple public datasets demonstrate that SEG-SAM not only outperforms SAM-based models in binary segmentation but also offers a significant advantage in semantic segmentation.
In conclusion, SEG-SAM establishes a new attempt for unified medical image segmentation. Despite its robust performance, challenges remain in handling ambiguous anatomical boundaries and minuscule structures.
In future work, we aim to extend this approach to medical video domains and refine the integration of language-driven semantics.

\section*{CRediT authorship contribution statement}
\textbf{Shuangping Huang:} Writing - original draft, Validation, Methodology.
\textbf{Hao Liang:} Writing - original draft, Software, Methodology.
\textbf{Qingfeng Wang:} Formal analysis, Investigation.
\textbf{Chulong Zhong:} Visualization, Data curation.
\textbf{Gang Wei:} Formal analysis, Supervision.
\textbf{Miaojing Shi:} Writing - review \& editing, Methodology.

\section*{Declaration of competing interest}
The authors declare that they have no known competing financial interests or personal relationships that could have appeared to influence the work reported in this paper.

\section*{Acknowledgments}
Supported by the National Key Research and Development Program of China (No.2023YFC3502900) and the National Natural Science Foundation of China (No.62576139, 62176093).












\bibliographystyle{cas-model2-names}

\bibliography{cas-refs}



\end{document}